\documentclass[sigconf]{acmart}

\usepackage[utf8]{inputenc}
\usepackage{kotex}
\usepackage[linesnumbered,ruled,vlined]{algorithm2e}
\usepackage{subfiles}
\usepackage{subcaption}
\usepackage{bm}

\newtheorem*{theorem*}{Theorem}
\newtheorem{proposition}{Proposition}
\newtheorem*{proposition*}{Proposition}

\newtheorem{corollary*}{Corollary}

\newtheorem*{lemma*}{Lemma}
\DeclareMathOperator*{\KL}{KL}

\newcommand{\E}{\mathbb{E}}
\newcommand{\N}{\mathcal{N}}
\newcommand{\bL}{\mathcal{L}}
\newcommand{\bb}[1]{\mathbf{#1}}
\newcommand{\bI}{\bb{I}}

\newcommand{\calN}{\mathcal{N}}
\usepackage{multirow}
\usepackage{capt-of}
\usepackage[]{graphicx}
\usepackage{graphicx}
\usepackage{adjustbox}
\usepackage{booktabs}
\usepackage{setspace}

\AtBeginDocument{%
	\providecommand\BibTeX{{%
			\normalfont B\kern-0.5em{\scshape i\kern-0.25em b}\kern-0.8em\TeX}}}

\copyrightyear{2022}
\acmYear{2022}
\setcopyright{acmcopyright}
\acmConference[KDD '22] {Proceedings of the 28th ACM SIGKDD Conference on Knowledge Discovery and Data Mining}{August 14--18, 2022}{Washington, DC, USA.}
\acmBooktitle{Proceedings of the 28th ACM SIGKDD Conference on Knowledge Discovery and Data Mining (KDD '22), August 14--18, 2022, Washington, DC, USA}
\acmPrice{15.00}
\acmISBN{978-1-4503-9385-0/22/08}
\acmDOI{10.1145/3534678.3539232}

\begin{document}
	
	\title{Learning Fair Representation via Distributional Contrastive Disentanglement}
	
	\author{Changdae Oh}
	\affiliation{%
		\institution{ University of Seoul }
		\city { Seoul}
		\country{Republic of Korea }}
	\email{ bnormal16@uos.ac.kr }
	
	\author{Heeji Won}
	\affiliation{%
		\institution{ Korea University }
		\city { Seoul}
		\country{Republic of Korea }}
	\email{ gmlwl1026@korea.ac.kr  }
	
	\author{Junhyuk So}
	\affiliation{%
		\institution{ POSTECH }
		\city { Pohang}
		\country{Republic of Korea }}
	\email{ junhyukso@postech.ac.kr }
	\author{Taero Kim}
	\affiliation{%
		\institution{ University of Seoul }
		\city { Seoul}
		\country{Republic of Korea }}
	\email{ rlaxofh123@uos.ac.kr }
	
	\author{Yewon Kim}
	\affiliation{%
		\institution{ University of Seoul }
		\city { Seoul}
		\country{Republic of Korea }}
	\email{ yeyewon12@uos.ac.kr }
	
	\author{Hosik Choi}
	\affiliation{%
		\institution{ University of Seoul }
		\city { Seoul}
		\country{Republic of Korea }}
	\email{ choi.hosik@uos.ac.kr }
	
	\author{Kyungwoo Song}
	\affiliation{%
		\institution{ University of Seoul }
		\city { Seoul}
		\country{Republic of Korea }}
	\email{ kyungwoo.song@uos.ac.kr }
	\authornote{Corresponding author.}

	\begin{abstract}
		Learning fair representation is crucial for achieving fairness or debiasing sensitive information. Most existing works rely on adversarial representation learning to inject some invariance into representation. However, adversarial learning methods are known to suffer from relatively unstable training, and this might harm the balance between fairness and predictiveness of representation. We propose a new approach, learning \textit{FAir Representation via distributional CONtrastive Variational AutoEncoder (FarconVAE)}, which induces the latent space to be disentangled into sensitive and non-sensitive parts. We first construct the pair of observations with different sensitive attributes but with the same labels. Then, FarconVAE enforces each non-sensitive latent to be closer, while sensitive latents to be far from each other and also far from the non-sensitive latent by contrasting their distributions. We provide a new type of contrastive loss motivated by Gaussian and Student-t kernels for distributional contrastive learning with theoretical analysis. Besides, we adopt a new swap-reconstruction loss to boost the disentanglement further. FarconVAE shows superior performance on fairness, pretrained model debiasing, and domain generalization tasks from various modalities, including tabular, image, and text.
	\end{abstract}


	
	\maketitle
	
	\section{Introduction}
	\label{s:introduction}
	Machine learning algorithms show the great success on many tasks, and they have been widely adopted in real-world applications. Most works of machine learning utilize a simple predictor at the test time, and the extracted features of a given dataset greatly influence the model prediction. So, the success of machine learning algorithms largely depends on the data representation that the model learned \cite{bengio2013representation}. However, the input features of a given dataset might contain noise and unnecessary information for the given task, and learning a representation that covers the important characteristics of the given dataset while invariant to unwanted information is crucial.
	
	Neural networks are known to have an advantage in representation learning. The learned representation from the neural net represents the characteristics of data with fewer dimensional vectors. Neural network based methods show a significant performance improvement on many domains including image \cite{he2016deep}, text \cite{kenton2019bert}, and tabular \citep{xu2019modeling}. However, recent studies raise that the traditional neural networks have difficulty in achieving fairness and domain generalization \cite{sarhan2020fairness, arjovsky2019invariant}. Neural networks provide a rich representation, but the representation also absorbs the sensitive (private) information or spurious correlation of a given dataset. Due to this unwanted information in learned representations, the model can induce unfair results in decision making system or fail to provide the correct predictions in the distribution shift scenario.
	
	There are many directions to remove the sensitive information and spurious correlation in the model. One of them, adversarial representation learning (ARL) methods \cite{xie2017controllable, roy2019mitigating} set two goals, (i) maximally retain salient information about a given target attribute, and (ii) minimize the information leakage about a given sensitive attribute. While these methods have shown compelling results when optimized successfully, their convergence instability hinders achieving the above goals and limits the wide use of ARL methods practically. Another promising direction is a disentangled representation learning \cite{creager2019flexibly, sarhan2020fairness} that separates the non-sensitive representation and sensitive representation. Their empirical success proves the effectiveness of disentanglement-based approaches to fair expression learning. However, most of the existing methods for disentangling \cite{kim2018disentangling, creager2019flexibly} also rely on the adversarial learning technique to approximate Total Correlation \cite{watanabe1960information} using the density ratio trick.
	
	The other direction is to learn the invariant representation of a given dataset. Invariant representation denotes the shared important features invariant across domains or environments. Recently, Arovsky et al. \cite{arjovsky2019invariant} proposed Invariant Risk Minimization (IRM) that encourages invariant representation learning with bi-level optimization. The representation trained with IRM may have shared important features, but there is no guarantee that the representation does not hold any sensitive information. Besides, Group-DRO \citep{sagawa2019distributionally} is known to be effective for learning robust representation, but it is also not guaranteed about the existence of sensitive information. Although invariant learning methods \citep{arjovsky2019invariant, creager2021environment, sagawa2019distributionally} show meaningful performance improvements on domain generalization tasks, we observe that the representation learned from IRM and Group-DRO still has a spurious correlation or sensitive information largely that poses a potential risk in real-world applications.
	
	In this paper, we propose FarconVAE (FAir Representation via distributional CONtrastive Variational AutoEncoder), a new disentangling approach with contrastive learning instead of adversarial learning. First, we construct a pair of two instances with different sensitive information and the same non-sensitive information. Then, FarconVAE 1) minimizes the distance between non-sensitive representations, 2) maximizes the dissimilarity between sensitive representations, and 3) maximizes the dissimilarity between sensitive and non-sensitive representations of each paired instance with our distributional contrastive loss. Finally, in the latent space divided into sensitive and non-sensitive representation, FarconVAE makes a fair prediction by using only non-sensitive representation.
	
	Besides, we adopt a new feature swap based objective for FarconVAE, swap-recon, to improve the disentanglement. Swap-recon replaces the non-sensitive representation of a given data instance with that of another paired data instance while leaving the sensitive representation untouched. By reconstructing from both the swapped latent and the original latent to the same original data instance, swap-recon further boosts the disentanglement. In Appendix \ref{a:results_ablation}, we empirically validate that swap-recon improves the representation disentanglement quality.
	
	To the best of our knowledge, this is the first study to improve both fairness and out-of-distribution generalization performance. FarconVAE provides a fair representation that only includes non-sensitive core information by disentangling the sensitive information. Besides, it also removes the spurious correlation in representation learned from IRM and Group-DRO, while maintaining the model accuracy. 
	
	Our contributions in this work are three-fold:
	\vspace{-0.5em}
	\begin{itemize}
		\item We propose a novel framework FarconVAE that learns disentangled invariant representation with contrastive loss to achieve algorithmic fairness and domain generalization.
		\item We provide a new distributional contrastive loss for disentanglement motivated by the Gaussian and Student-t kernels.
		\item The proposed method is theoretically analyzed and empirically demonstrated on a broad range of data types (tabular, image, and text) and tasks, including fairness and domain generalization.
	\end{itemize}
	
	\section{Related Works}
	\label{s:relatedworks}
	\subsection{Domain Generalization}
	The primary objective of domain generalization is to show stable predictive performance even though the test distributions are different from the train distribution \cite{gulrajani2020search}. Traditional machine learning algorithm fails domain generalization, and there have been many research works to handle the distribution shift in diverse ways, including Bayesian neural net \cite{maddox2019simple}, data augmentation \cite{xu2020adversarial}, robust optimization \cite{sagawa2019distributionally, levy2020large}, and invariant learning \cite{arjovsky2019invariant, creager2021environment}. 
	
	Recently, Group-Distributionally Robust Optimization (Group-DRO) \cite{sagawa2019distributionally} and Invariant Risk Minmimzation (IRM) \cite{arjovsky2019invariant} have been considered promising ways for domain generalization.
	Group-DRO optimizes the models to focus on the worst-case group with strong regularization, and IRM encourages the model to focus on the shared essential features across the group. However, there is no theoretical guarantee that the Group-DRO and IRM alleviate the spurious correlation or sensitive information removal. Recent works point out that the performance improvements of Group-DRO and IRM are limited on a shifted test distribution \cite{gulrajani2020search}.
	
	\subsection{Learning Fair Representation}
	Fair representation learning aims to learn a representation that can be used for making accurate predictions without bias from sensitive information. We can divide the various fairness works depending on whether adversarial learning is used, and we introduce the related works for fair representation learning in this subsection.
	\subsubsection{Fair Representation with Adversarial Learning}
	Generative adversarial learning (GAN) \cite{goodfellow2014generative} shows a significant improvement in density estimation, and it has been widely utilized in many tasks. For fair representation learning, density estimation of a given instance without sensitive information is necessary. Therefore, there have been many research works that adopt adversarial learning for fair representation \cite{xie2017controllable, madras2018learning, zhang2018mitigating, roy2019mitigating}. Controllable Invariance (CI) \cite{xie2017controllable} adopts adversarial min-max game to filter out detrimental features such as sensitive information. CI introduces three types of network; encoder, discriminator, predictor, and adversarial minimax game between encoder and discriminator encourages the representation is invariant to sensitive information. Maximum Entropy Adversarial Representation Learning (MaxEnt-ARL) \cite{roy2019mitigating} is another kind of adversarial method, and MaxEnt-ARL utilizes a non-zero-sum game adversarial formulation by adopting different objectives for generator and discriminator to overcome the sub-optimal problems. The order approach is Flexibly Fair Variational AutoEncoder (FFVAE) \cite{creager2019flexibly} based on disentangled representation learning. FFVAE learns the separated latent space for sensitive and non-sensitive information. Although the algorithm does not contain an explicit adversary, it also relies on adversarial learning to approximate the Total Correlation (TC) \cite{watanabe1960information} penalty used for disentanglement.
	
	However, adversarial learning is known to have convergence instability problems, and it might hinder learning the robust representation. We empirically observe that the previous adversarial learning-based fair algorithm has difficulty separating the sensitive information. To solve the problems, we propose a contrastive learning-based disentangled representation learning method, FarconVAE, instead of adversarial learning.
	\subsubsection{Fair Representation without Adversarial Learning}
	Recently, there have been other fair representation learning research works without adversarial learning \cite{zemel2013learning, louizos2015variational, cheng2020fairfil, sarhan2020fairness}. Zemel et al. \cite{zemel2013learning} propose fair clustering methods with probabilistic mapping, and Variational Fair AutoEncoder (VFAE) \cite{louizos2015variational} adopts Maximum Mean Discrepancy (MMD) measure \cite{gretton2006kernel} to penalize the posterior. Besides, Fair Filter (FairFil) \cite{cheng2020fairfil} removes the sensitive information that is inherent in the sentence embedding from a pretrained language model. To learn debiased embeddings, FairFil utilizes contrastive learning and mutual information estimator.
	
	Orthogonal Disentangled Fair Representations (ODFR) \cite{sarhan2020fairness} is a non-adversarial disentangle-based methods, and it introduces orthogonal priors to enforce an orthogonality constraint between sensitive and non-sensitive representation. $q_{\phi_{s}}(z_{s}|x)$ and $q_{\phi_{x}}(z_{x}|x)$ denote the posteriors of sensitive and non-sensitive representation parameterized by $\phi_{s}$ and $\phi_{x}$ respectively, and $p(z_{x})$ and $p(z_{s})$ denote the priors, where $p(z_{x})=\calN([1,0]^T,\bI)$ and $p(z_{s})=\calN([0,1]^T,\bI)$. 
	\begin{equation}
		L_{OD} = \KL((q_{\phi_{x}}(z_{x}|x)||p(z_{x})) + \KL((q_{\phi_{s}}(z_{s}|x)||p(z_{s}))
		\label{eq:odfr}
	\end{equation}
	
	However, minimizing $L_{OD}$ in Eq. \ref{eq:odfr} does not guarantee a robust fair representation. It is known that the vanilla KL-divergence between posterior and fixed prior might cause the posterior collapse, and the representation might not contain meaningful information of given data \cite{he2018lagging}. As a result, ODFR adopts additional auxiliary components such as entropy loss. Besides, there are diverse ways to set orthogonal priors, and the performance of ODFR may largely depend on the choice of prior.
	
	FarconVAE has a relationship with FairFil and ODFR, but there are three major differences. First, FarconVAE measures the divergence between two posteriors for disentanglement instead of the fixed prior in ODFR. Thus, FarconVAE is relatively free from posterior collapse problems and prior choice. Second, we adopt a new distributional contrastive loss motivated by Gaussian and Student-t kernel. It makes FarconVAE get highly disentangled fair representation without the need for auxiliary components such as discriminators and entropy loss. Third, in contrast to ODFR and FairFil, which have shown effectiveness in limited data types for only fair prediction tasks, our FarconVAE has validated on both fairness and domain generalization tasks across three representative data types.
	
	\subsection{Contrastive Learning}
	Recently, Contrastive learning has emerged as a new promising paradigm for self-supervised representation learning, showing powerful performance in a broad domain such as Vision \cite{chen2020simple}, NLP \cite{gao2021simcse}, and Graph \cite{li2019graph}. InfoNCE \cite{oord2019representation} is a widely used loss function for contrastive learning that effectively formulates the instance discrimination task. InfoNCE loss encourages the similarity between positive pairs and the dissimilarity between negative pairs. Representation space learned by contrastive learning has performed well in various downstream tasks such as classification, clustering, or semantic similarity evaluation. However, there is limited work \cite{weinberger2022moment} that handled the disentanglement with contrastive learning. This paper provides a new distributional contrastive learning method to obtain disentangled invariant representation and apply it to fairness and domain generalization tasks. 
	
	\section{Methodology}
	\label{s:methodology}
	In this section, we firstly introduce the basic notation and problem formulation in Section \ref{s:methodology}. In Sections \ref{s:methodology_farconvae}, \ref{s:methodology_contra}, we describe FarconVAE in detail, and cover the kernel motivated distributional contrastive loss that induces stable disentanglement, respectively. In Section \ref{s:methodology_swap}, we provide a description for swap-recon, a new feature swap based regularization. Finally, we theoretically analyze our two specified contrastive losses in Section \ref{s:methodology_analysis}.
	
	\begin{figure}
		\centering
		\begin{minipage}{.45\linewidth}
			\begin{subfigure}[b]{.9\linewidth}
				\includegraphics[width=\textwidth]{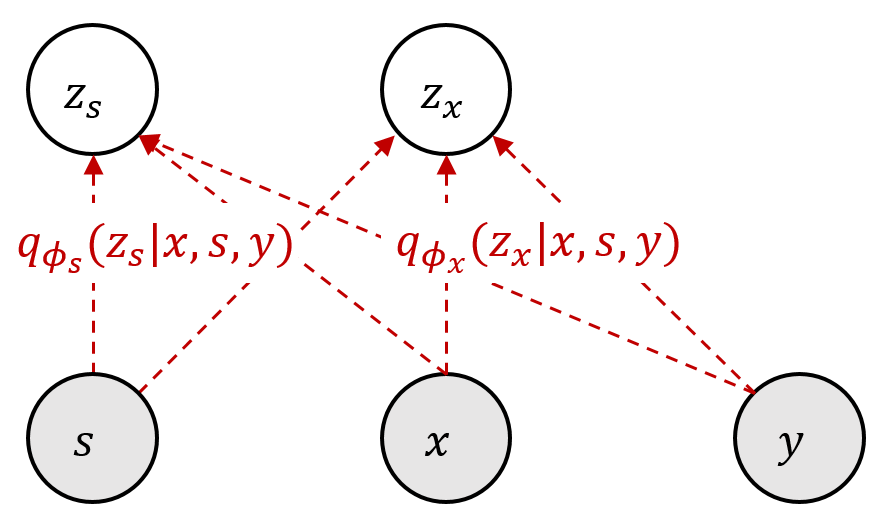}
				\caption{Recognition model}
				\label{fig:pgm_farconvae_infer}
			\end{subfigure} 
		\end{minipage}
		\begin{minipage}{.45\linewidth}
			\begin{subfigure}[t]{.9\linewidth}
				\includegraphics[width=\textwidth]{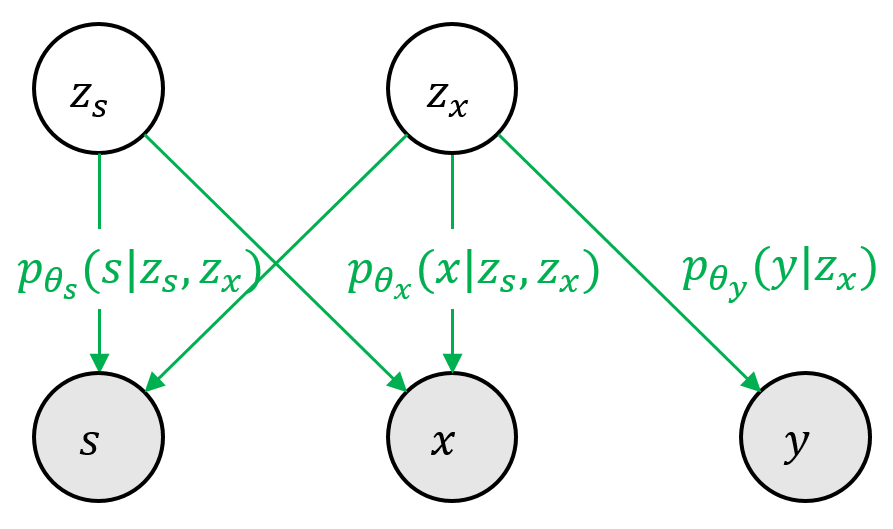}
				\caption{Generative model}
				\label{fig:pgm_farconvae_gen}
			\end{subfigure}
		\end{minipage}
		\begin{minipage}{.80\linewidth}
			\begin{subfigure}[t]{.9\linewidth}
				\includegraphics[width=\textwidth]{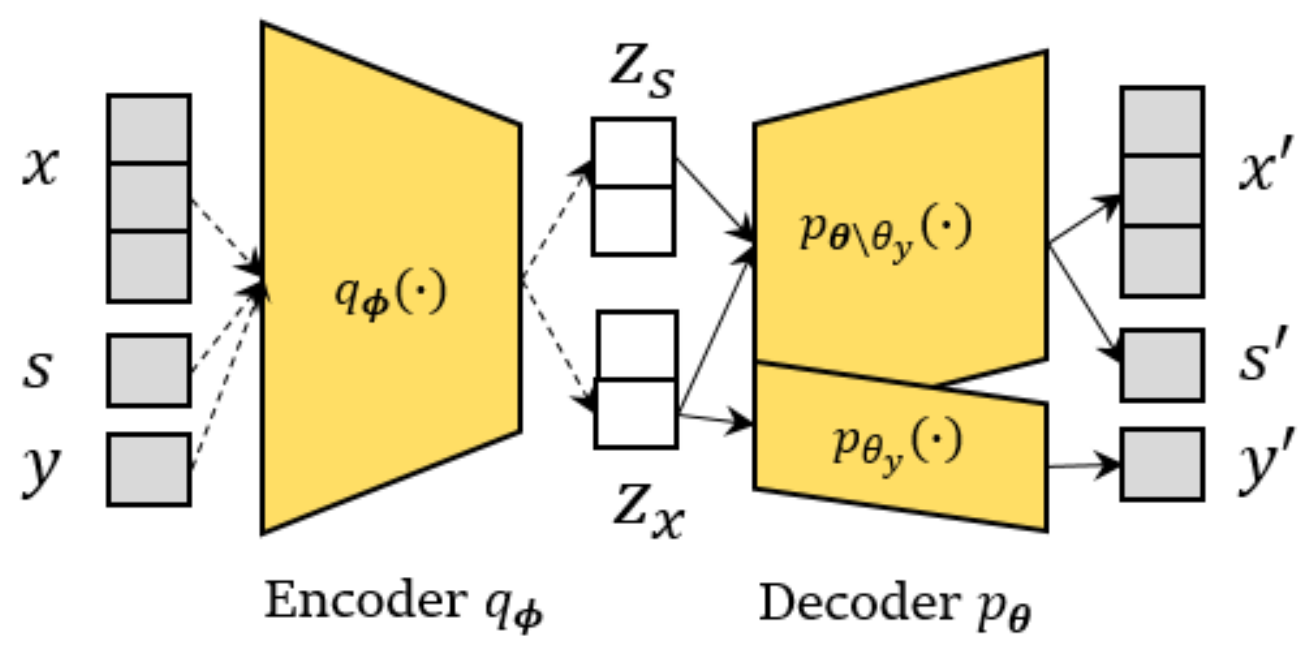}
				\caption{FarconVAE in Neural Networks (NN) view}
				\label{fig:nn_farconvae}
			\end{subfigure}
		\end{minipage}
		\caption{Graphical notations of FarconVAE (a), (b), and neural network view (c). The dashed line denotes the recognition models, neural network with parameter $\phi$, and the solid line represents the generative models, neural network with parameter $\theta$. 
			Also, the gray color denotes the observed variables, while the white color denotes the latent variables. 
		}
		\label{fig:FarconVAE_pgm_nn}
	\end{figure}
	
	\subsection{Problem Definition}
	\label{s:methodology}
	Let $x$ be an observed input feature, $s$ be its sensitive attribute, and $y$ be a target label. As described in \cite{arjovsky2019invariant, sarhan2020fairness}, the sensitive attribute $s$ is highly correlated with feature $x$ and spuriously correlated with label $y$ on many real-world datasets, even if it is essentially irrelevant. In this situation, We want to build a fair ML algorithm. Mathematically, fairness can be defined as $p_{\theta}(y|x)=p_{\theta}(y|x,s)$ \cite{sarhan2020fairness}. The Fairness definition shows that the model output needs to be independent of the sensitive attribute. As a result, a fair representation debiased w.r.t sensitive information is necessary to achieve fairness. Furthermore, performance degradation for target label prediction should be minimized. However, the existing methods for building fair algorithms have difficulty in learning robust representation satisfying both predictiveness and fairness, as stated in Section \ref{s:relatedworks}.
	
	In this work, we propose FarconVAE, which aims to learn a function that maps each data $(x,s,y)$ to disentangled representation $z_{s}$ and $z_{x}$ on the two separated latent spaces $\mathcal{Z}_{s}, \mathcal{Z}_{x}$ that involve different semantic meanings. Specifically, $\mathcal{Z}_{s}$ is a latent space to absorb and isolate the whole sensitive information from observation, and $\mathcal{Z}_{x}$ maximally preserves only non-sensitive information to predict the target accurately. In other words, $z_{x}$ is a fair representation used for target prediction that has high predictiveness without relying on sensitive information.
	
	\begin{figure*}[t!] 
		\centerline{\includegraphics[width=0.65\textwidth]{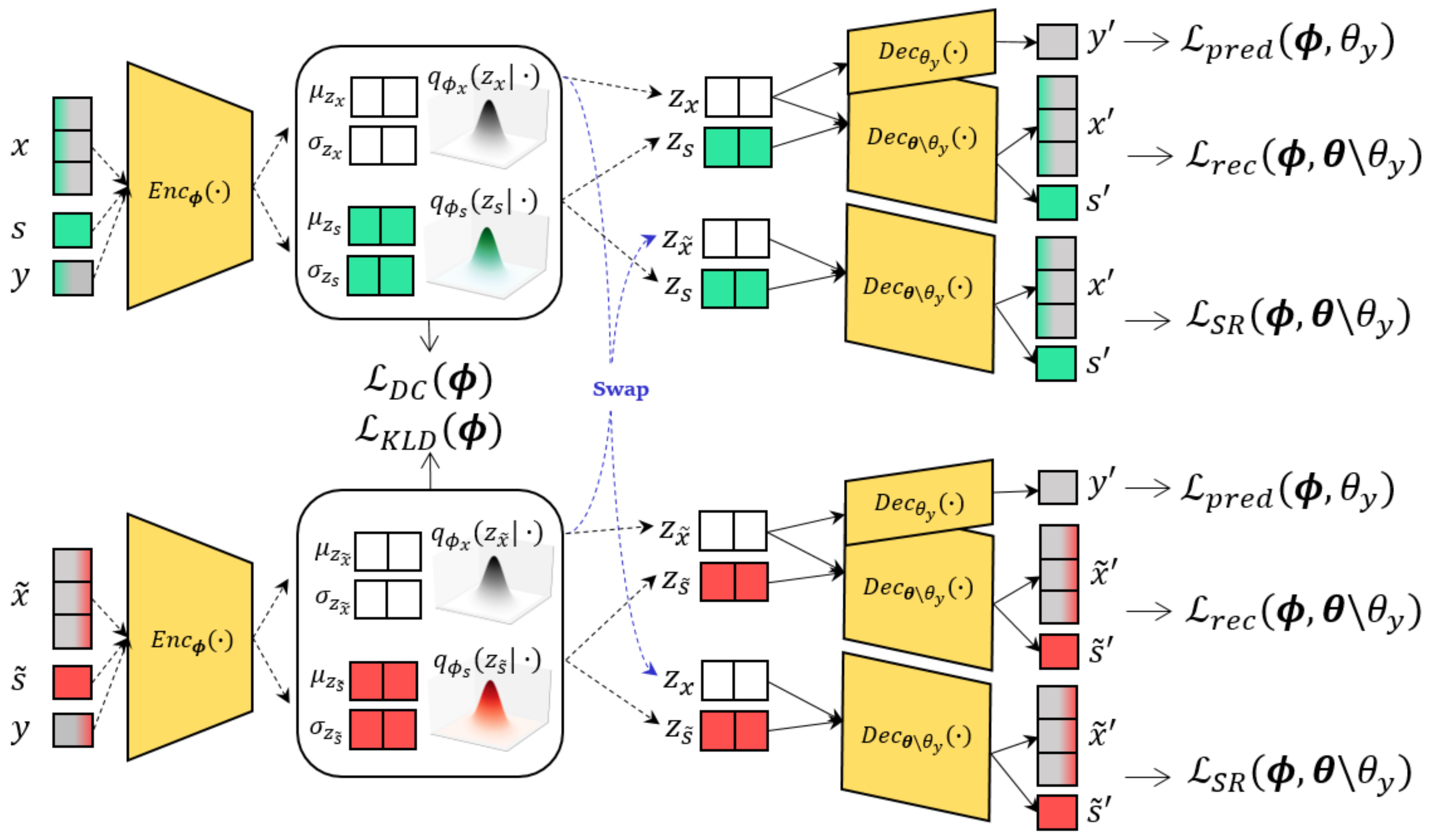}}
		\caption{Overall framework. Green and Red colors denote the sensitive information, while white denotes non-sensitive residual information. When the original data $(x,s,y)$ is given, we find or generate the data $(\tilde{x},\tilde{s},y)$ that has a different sensitive attribute while having the same target label for constructing pair. After that, we extract latent representation $z_{x}, z_{s}, z_{\tilde{x}}, z_{\tilde{s}}$. For given representations, we enforce 1) similarity between $z_{x}$ and $z_{\tilde{x}}$, 2) dissimilarity between $z_{s}$ and $z_{\tilde{s}}$ 3) dissimilarity between $z_{x}$ (or $z_{\tilde{x}}$) and $z_{s}$ (or $z_{\tilde{s}}$) using distribution contastive loss for disentangling. Moreover, we adopt swap-recon (SR) that further enhances the disentanglement by performing consistent reconstruction with both swapped and original latent. As a result, FarconVAE provides the disentangled fair representation $z_{x}$. In training, we put $y$ as input to FarconVAE for better predictiveness, but it requires the true label in test time. We solve this by using the best classifier that predicts $y$ from $x$. See Appendix \ref{a:implementation} for details.}
		\label{fig:FarconVAE_framework}
	\end{figure*}
	
	\subsection{FarconVAE Structure}
	\label{s:methodology_farconvae}
	Figure \ref{fig:FarconVAE_pgm_nn} represents the FarconVAE in terms of graphical model and neural net view.  FarconVAE assumes that there are three observable variables, input feature $x$, sensitive attribute $s$, and target label $y$. FarconVAE has one encoder $p_{\bm\phi}(\cdot)$ that splits into two heads for encoding $z_x$ and $z_s$, and two decoders: $p_{\theta{y}}(\cdot)$ for decoding $y$ and $p_{\bm\theta \textbackslash \theta{y}}(\cdot)$ that splits into two heads for decoding $x$ and $s$. The goal of FarconVAE is to capture two disentangled latent $z_{s}$ and $z_{x}$, where $z_{s}$ and $z_{x}$ denote the sensitive and non-sensitive representation, respectively. With these two latent variables, The log marginal likelihood of the $x,s$ and $y$ with model parameters $\bm{\theta}$ is as follows:
	\begin{equation}
		\log{p_{\bm{\theta}}(x,s,y)} = \log\int\int{p_{\bm{\theta}}(x,s,y,z_{x},z_{s})dz_{x}dz_{s}} \label{eq:FarconVAE_marginal}\\
	\end{equation}
	\subsubsection{Model Inference} Direct optimization of marginal likelihood, Eq. \ref{eq:FarconVAE_marginal}, is intractable, so we utilize the variational inference \cite{jordan1999introduction} to approximate the marginal likelihood. We introduce the variational distribution $q$, and we adopt encoder $q_{\bm{\phi}}$ and decoder $p_{\bm{\theta}}$ as neural networks parameterized by $\bm{\phi}=\{\phi_{body},\phi_{head_x},\phi_{head_s}\}$ and $\bm{\theta}=\{\theta_{body},\theta_{head_x},\theta_{head_s},\theta_{y}\}$, respectively. In the following all sections, we will notate $\{\phi_{body} \cup \phi_{head_x}\}$ as $\phi_{x}$ and $\{\phi_{body} \cup \phi_{head_s}\}$ as $\phi_{s}$, $\{\theta_{body} \cup \theta_{head_x}\}$ as $\theta_{x}$, and $\{\theta_{body} \cup \theta_{head_s}\}$ as $\theta_{s}$.
	
	For the disentanglement between $z_{x}$ and $z_{s}$, the conditional independence between $z_{x}$ and $z_{s}$ given $x,s,y$ is necessary. Under the conditional independence assumption, we construct the variational objective \cite{jordan1999introduction,DBLP:journals/corr/KingmaW13}, called evidence lower bound (ELBO), as follows:
	\begin{flalign}
		& \log{p_{\bm{\theta}}(x,s,y)} \nonumber &\\
		& \geq \E_{q_{\bm{\phi}}}[\log p_{\theta_{x}}(x|z_{x},z_{s}) + 
		\log p_{\theta_{s}}(s|z_{x},z_{s}) + \log p_{\theta_{y}}(y|z_{x})] \nonumber &\\
		& -\KL(q_{\phi_{x}}(z_{x}|x,s,y)||p(z_{x})) -\KL(q_{\phi_{s}}(z_{s}|x,s,y)||p(z_{s})) &\label{eq:KLD} \\ 
		& = \bL_{ELBO}(\bm{\phi},\bm{\theta};x,s,y) & \label{eq:FarconVAE_ELBO}
	\end{flalign}
	$\bL_{ELBO}$ in Eq. \ref{eq:FarconVAE_ELBO} consists of three components. First, the KL divergence in Eq. \ref{eq:KLD} can be interpreted as a regularization term to prevent the variational posteriors $q_{\phi_{x}}(z_x|x,s,y)$ and $q_{\phi_{s}}(z_s|x,s,y)$ moving too far from their priors. Second, reconstruction terms, $p_{\theta_{x}}(x|z_x,z_s)$ and $p_{\theta_{s}}(s|z_x,z_s)$ encourage the latent representation $z_{x}$ and $z_{s}$ to preserve the salient information of $x$ and $s$. Third, prediction term $p_{\theta_{y}}(y|z_x)$ gives an ability to model to predict the target value, and it injects task-specific information to representation $z_{x}$.
	
	By maximizing the ELBO in Eq. \ref{eq:FarconVAE_ELBO}, we can infer the parameters of distribution over the joint latent variables $z_s$ and $z_x$ that depend on $x,s$, and $y$. These two latent representations include the salient information of the given dataset. Different with $z_{s}$, $z_{x}$ has capability to predict $y$, and its difference contributes to disentanglement between $z_{s}$ and $z_{x}$.
	However, FarconVAE can be improved in two ways. First, $\bL_{ELBO}$ just assumes the conditional independence between $z_{x}$ and $z_{s}$ given $x,s,y$. Second, $z_{x}$ still has sensitive information if the $y$ is spuriously correlated with $s$. Therefore, we provide a new contrastive loss with FarconVAE and introduce it in Section $\ref{s:methodology_contra}$.
	
	\subsubsection{Model in Detail} In this subsection, we provide the distributional assumption in FarconVAE. For the prior $p(z_{s})$ and $p(z_{x})$, we adopt standard Gaussian $\N(z_{s};0,\bI)$ and $\N(z_{x};0,\bI)$, respectively.
	
	For encoder, we formulate the variational posterior distribution $q_{\bm{\phi}}(\cdot|\cdot)$ as Gaussian distribution with diagonal covariance structure, parameterized by neural network. In the remaining sections, we will omit the parameter of posterior and refer to it as $q(\cdot|\cdot)$. In the formulas below, $z_{\cdot}$ can be $z_{x}$ or $z_{s}$.
	\begin{align}
		\log q_{\phi_{\cdot}}(z_{\cdot}|x,s,y) = \log \N (z_{\cdot};\mu_{z_{\cdot}},\sigma^2_{z_{\cdot}} \bI) \nonumber
	\end{align}
	
	For decoder, we adopt Gaussian distribution for continuous variables and Bernoulli distribution for discrete variables. Like above, the parameters of the distribution are parameterized by $\bm\theta$.
	\begin{align}
		p_{\theta_{x}}(x|z_{x}, z_{s}) = \N (x;\mu_{x},\sigma^2_{x}\bI) \; \text{or} \; p_{\theta_{x}}(x|z_{x}, z_{s}) = \text{Ber}(x;\mu_{x}) \nonumber\\
		p_{\theta_{s}}(s|z_{x}, z_{s}) = \N (s;\mu_{s},\sigma^2_{s}\bI) \; \text{or} \; p_{\theta_{s}}(s|z_{x}, z_{s}) = \text{Ber}(s;\mu_{s}) \nonumber \\
		p_{\theta_{y}}(y|z_{x}) = \N (y;\mu_{y},\sigma^2_{y}\bI) \; \text{or} \; p_{\theta_{y}}(y|z_{x}) = \text{Ber}(y;\mu_{y})\nonumber
	\end{align}
	To reconstruct feature $x$ and sensitive attribute $s$, FarconVAE uses both $z_{x}$ and $z_{s}$, while for predicting target label $y$ it utilizes $z_{x}$ only.
	
	\subsection{Contrastive Learning for Disentanglement}
	\label{s:methodology_contra}
	For given $x, s, y$, we can estimate the counterfactual that has the same label $y$, but different sensitive attribute $\tilde{s}$ and its corresponding $\tilde{x}$ (similar to $x$). Counterfactual estimation can be replacing the given sensitive $s$, as opposite $s'$ \cite{kim2021counterfactual}, or finding the similar instance that has different sensitive attribute \cite{shalit2017estimating}. Given original data instance and its counterfactual, we train the model with a pair of them with $\bL_{ELBO}$ in Eq. \ref{eq:FarconVAE_ELBO}. Besides, we introduce the contrastive loss to ensure the disentanglement between $z_{x}$ and $z_{s}$. For the latent representation of given dataset and its counterfactual, $z_{x}, z_{s}, z_{\tilde{x}}, z_{\tilde{s}}$, the similarity between $z_{x}$ an $z_{\tilde{x}}$, the difference between $z_{s}$ and $z_{\tilde{s}}$, and the difference between $z_{x}$ (or $z_{\tilde{x}}$) and $z_{s}$ (or $z_{\tilde{s}}$) are required.
	
	In this perspective, we regard $(z_x, z_{\tilde x})$ as a positive pair and $(z_s, z_{\tilde s})$, $(z_x, z_{s})$, $(z_{\tilde{x}}, z_{\tilde s})$ as negative pairs by analogy with problem setting in contrastive learning literature. To encourage the similarity between positive representation pair and the difference between negative representation pair, we \textit{minimize the divergence} between posterior $(q(z_x|\cdot), q(z_{\tilde{x}}|\cdot))$ and \textit{minimize the similarity} between posterior $(q(z_s|\cdot), q(z_{\tilde s}|\cdot))$, $(q(z_x|\cdot), q(z_{s}|\cdot))$, and $(q(z_{\tilde{x}}|\cdot), q(z_{\tilde{s}}|\cdot))$ simultaneously using \textit{Distributional Contrastive} loss:
	
	\begin{align}
		\begin{split}
			\bL_{DC} & = \mathcal{D}_{z}(q(z_x|\cdot), q(z_{\tilde{x}}|\cdot)) + \sum_{neg \in \textbf{NPP}}\textit{k}(\mathcal{D}_{z}(neg)) \label{eq:s_contra}
		\end{split}
	\end{align}
	where $\mathcal{D}_{z}$ is an arbitrary divergence between two distributions, $k(\cdot)$ is any positive definite kernel function (e.g. Gaussian RBF), and $\textbf{NPP} = \{(q(z_s|\cdot), q(z_{\tilde{s}}|\cdot)), (q(z_x|\cdot), q(z_s|\cdot)), (q(z_{\tilde{x}}|\cdot), q(z_{\tilde{s}}|\cdot))\}$ is a set of \textbf{n}egative \textbf{p}osterior \textbf{p}airs. Thus, $\mathcal{D}_{z}(neg)$ denotes a divergence between two negatively paired distributions. For convenience in implementation, we choose $\mathcal{D}_{z}$ as the averaged KL divergence that simply averages two possible KL divergences between probability distributions, i.e., $\Bar{\KL}(p_{0}, p_{1})={ (KL(p_{0}||p_{1})+KL(p_{1}||p_{0})) \over 2}$. Note that, the divergence is both the statistical distance and the generalized squared distance between two distributions. Consequently, after being passed through the kernel function $\textit{k}(\cdot)$, the distance between two distributions is converted into the similarity between two distributions.
	
	By adopting Gaussian or Student-t kernel for kernel function $\textit{k}(\cdot)$, we specify our two \textit{Distributional Contrastive} losses as below:
	\begin{align}
		\bL_{DC-t} &= \Bar{\KL}(q(z_x|\cdot)||q(z_{\tilde x}|\cdot)) + \sum_{neg \in \textbf{NPP}}(1 + \Bar{\KL} (neg))^{-1} \label{eq:student-t}\\
		\bL_{DC-G} &= \Bar{\KL}(q(z_x|\cdot)||q(z_{\tilde x}|\cdot)) + \sum_{neg \in \textbf{NPP}}\exp( - \Bar{\KL} (neg)) \label{eq:gaussian}
	\end{align}
	In Section \ref{s:methodology_analysis}, we present a theoretical analysis for these losses.
	
	\subsection{Swap-Reconstruction}
	\label{s:methodology_swap}
	Besides distributional contrasting, we perform the swap-reconstruction task (referred to \textit{swap-recon}) that improves disentanglement further. After passing the encoder, the observation pair $\{(x,s,y), (\tilde x,\tilde s,y)\}$ is encoded as $\{(z_x,z_s)$, $(z_{\tilde x}, z_{\tilde s})\}$. As mentioned above subsection, we hope that only $z_s$ and $z_{\tilde s}$ contain sensitive information and $z_x$ and $z_{\tilde x}$ have non-sensitive residual information. For taking a step closer to this goal, we swap the non-sensitive representations $z_x$ and $z_{\tilde x}$ each other and form new concatenated latent vector pairs $(z_{\tilde x}, z_{s})$ and  $(z_{x}, z_{\tilde s})$. To reconstruct the original observation $(x,s,y)$ and $(\tilde x,\tilde s,y)$ from the swapped latent $(z_{\tilde x}, z_{s})$ and  $(z_{x}, z_{\tilde s})$, FarconVAE is incentivized to encode only non-sensitive information in $z_x$ and $z_{\tilde{x}}$.
	In the early training stage, the swap-recon generates diverse combinations of non-sensitive and sensitive information, and we observe that it improves not only disentanglement, but also generalization performance. The loss term is defined as:
	\begin{align}
		\bL_{SR} & = {1 \over 2}((\log p_{\theta_{x}}(x|z_{\tilde x},z_{s}) + \log p_{\theta_{s}}(s|z_{\tilde x},z_{s})) \nonumber \\
		& + ( \log p_{\theta_{x}}(\tilde x|z_{x},z_{\tilde{s}}) + \log p_{\theta_{s}}(\tilde s|z_{x},z_{\tilde{s}}))) \label{eq:L_SR}
	\end{align}
	There were similar methods to our swap-recon, and they showed promising results in disentangling \cite{mathieu2016disentangling} or debiasing \cite{kim2021biaswap}. 
	However, this is the first study that proposes swap-recon under the contrastive learning framework. We utilize $\bL_{SR}$, swap-recon loss, with our main distributional contrastive loss $\bL_{DC}$.
	
	\subsubsection{Overall Objective}
	FarconVAE is a general framework for disentangled fair representation learning. To encourage disentanglement, we use a distributional contrastive loss and swap-recon loss as shown in Figure \ref{fig:FarconVAE_framework}. The overall objective becomes as follows:
	\begin{align}
		& \bL_{Farcon}(\bm\phi,\bm\theta;x,\tilde x,s,\tilde s,y) \nonumber \\
		& = {1 \over 2}(\bL_{ELBO}(\bm\phi,\bm\theta;x,s,y) + \bL_{ELBO}(\bm\phi,\bm\theta;\tilde x,\tilde s,y)) \nonumber \\
		& + \alpha \bL_{DC}(\bm\phi;x,\tilde x,s,\tilde s,y) + \gamma \bL_{SR}(\bm\phi,\bm\theta\textbackslash\theta_{y};x,\tilde x,s,\tilde s,y) \nonumber
	\end{align}
	where $\alpha$ and $\gamma$ are hyperparameters to weigh the distributional contrastive loss and the swap-recon loss, respectively. $\bL_{ELBO}$ term is optimized over the given original data instance $(x, s, y)$, and its counterfactual instance $(\tilde{x}, \tilde{s}, y)$ individually. On the other hand, $\bL_{DC}$ and $\bL_{SR}$ terms are optimized over both original and counterfactual instances simultaneously. We demonstrate the effectiveness of proposed loss terms $\bL_{DC}$ and $\bL_{SR}$ in Appendix \ref{a:results_ablation}.
	\begin{figure}[h!]
		\begin{subfigure}{0.23\textwidth}
			\includegraphics[width=0.9\textwidth]{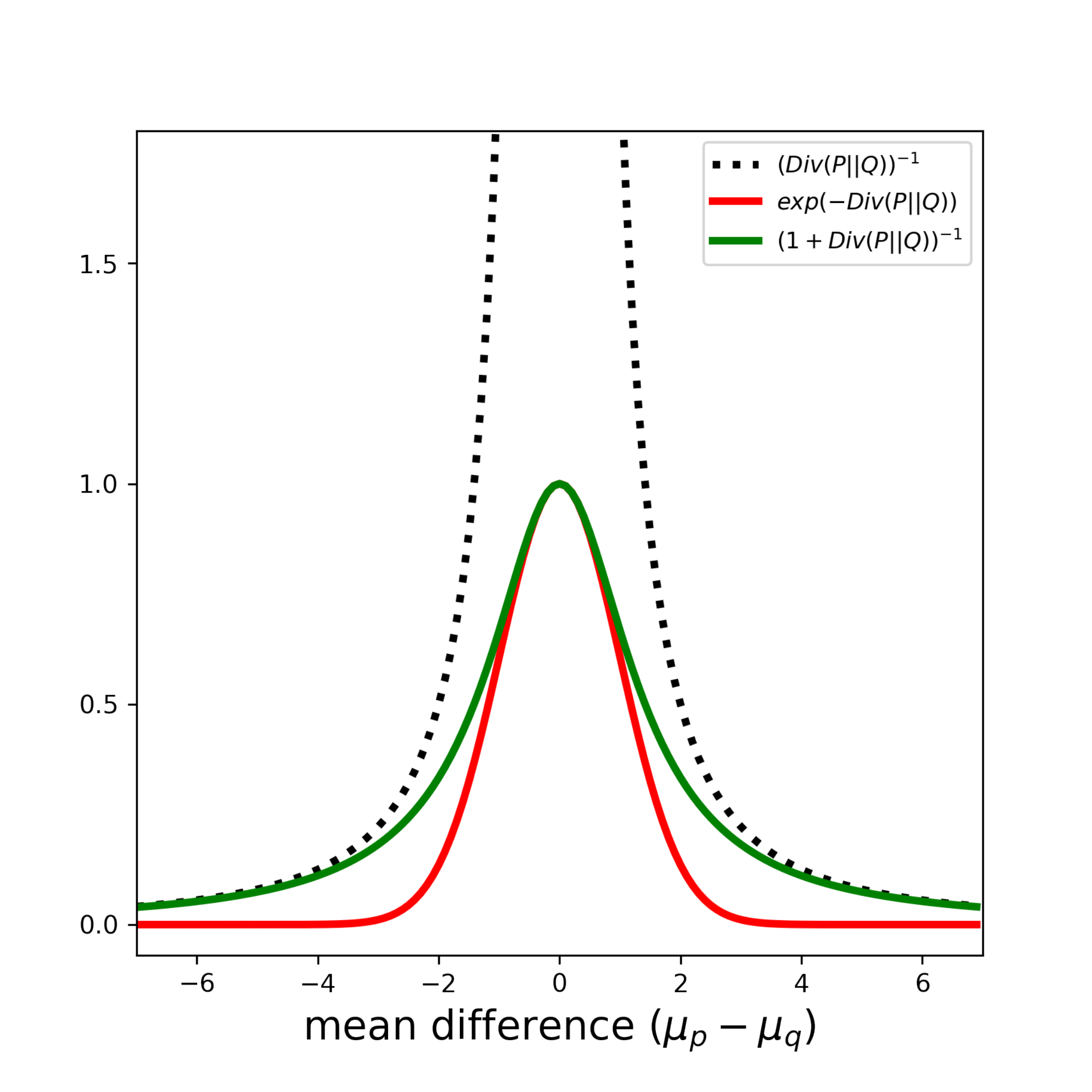}
			\caption{Mean difference}
		\end{subfigure}
		\begin{subfigure}{0.23\textwidth}
			\includegraphics[width=0.9\textwidth]{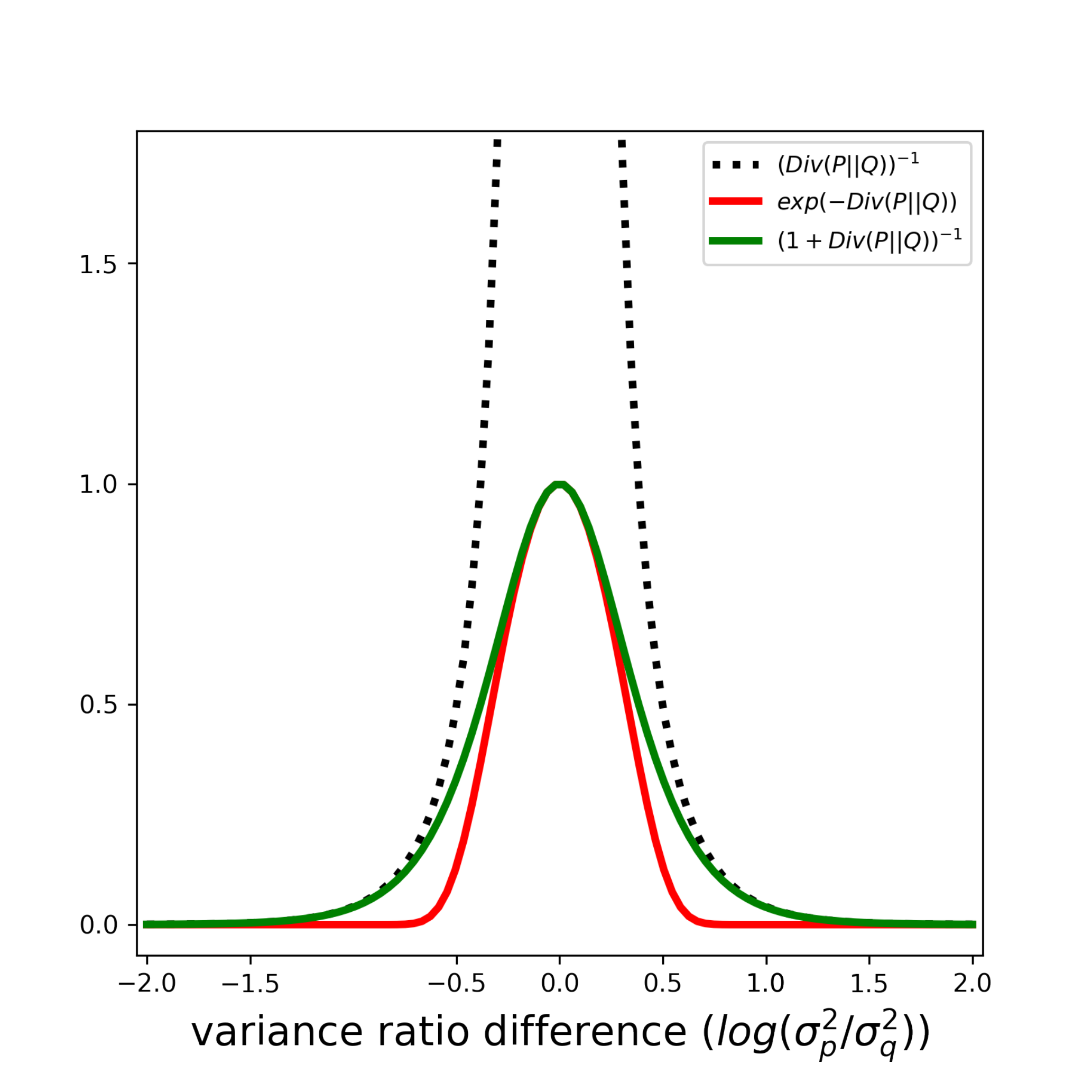}
			\caption{Variance difference}
		\end{subfigure}
		\caption{Comparison among similarities. In (a), we set the same variance and vary the means. In (b), we set the same mean and vary the variances.}
		\label{fig:divergence_change}
	\end{figure}
	\begin{table*}[t!]
		\caption{Fair classification results on the Adult, German, and YaleB datasets. 
			Random-Guess denotes the $y$ and $s$ accuracy when we predict the class randomly. Therefore, the $s$ accuracy of Random-Guess is the barometer to determine whether the representation has sensitive information or not. 
			$^{*}$ denotes the reported performance in the supplementary material of \cite{sarhan2020fairness}.}
		\label{tab:fair_classification}
		\begin{tabular}{l|cc|cc|cc}
			\toprule
			& \multicolumn{2}{c|}{Adult} & \multicolumn{2}{c|}{German} & \multicolumn{2}{c}{YaleB} \\
			\midrule
			Model & $y$ accuracy  & $s$ accuracy & $y$ accuracy & $s$ accuracy & $y$ accuracy & $s$ accuracy \\
			\midrule
			Random-Guess & 50.00 & 50.00 & 50.00 & 50.00 & 2.63 & 20.00 \\
			\midrule
			LFR* & 82.30 & \underline{67.00} & 72.30 & 80.50 & - & - \\
			VAE* & 81.90 & \textbf{66.00} & 72.50 & 79.50 & - & - \\
			VFAE*& 81.30 & \underline{67.00} & 72.70 & 79.70 & 85.00 & 57.00 \\
			CI & 84.49$\pm{0.15}$ & 68.62$\pm{7.19}$ 
			& 80.05$\pm{0.16}$ & 51.70$\pm{18.16}$ 
			& 89.79$\pm{0.33}$ & 8.67$\pm{1.83}$ \\
			MaxEnt-ARL & 84.52$\pm{0.06}$ & 67.90$\pm{0.94}$ 
			& 78.75$\pm{0.35}$ & \underline{49.25$\pm{8.82}$}
			& 89.95$\pm{0.30}$ & 8.67$\pm{1.83}$\\
			ODFR & 84.43$\pm{0.11}$ & 68.93$\pm{0.15}$ 
			& 78.05$\pm{1.34}$ & 53.60$\pm{12.86}$ 
			& 85.09$\pm{0.86}$ & 14.84$\pm{17.91}$\\
			FarconVAE-t & \underline{84.67$\pm{0.02}$} & 67.38$\pm{0.00}$
			& \textbf{83.35$\pm{0.75}$} & 54.24$\pm{18.31}$ 
			& \underline{90.96$\pm{0.18}$} & \textbf{20.01$\pm{8.30}$}\\
			FarconVAE-G & \textbf{84.67$\pm{0.01}$} & 67.36$\pm{0.04}$ 
			& \underline{83.20$\pm{0.35}$} & \textbf{49.80$\pm{13.38}$}
			& \textbf{90.99$\pm{0.04}$} & \underline{18.07$\pm{13.27}$}\\
			\bottomrule
		\end{tabular}
	\end{table*}
	\subsection{Theoretical Analysis}
	\label{s:methodology_analysis}
	\label{s:method4}
	This subsection presents the theoretical analysis of our distributional contrastive loss. The naive way to achieve contrastive disentangling is to use a simple reciprocal of divergence as a $k(\cdot)$ in Eq. \ref{eq:s_contra}. However, setting $k(\cdot)$ as a simple reciprocal might cause a numeric instability, as shown in Figure \ref{fig:divergence_change}. To mitigate the numerical instability and to encourage disentanglement, we provide new kernel motivated contrastive learning. We formulate Gaussian Kernel motivated similarity $exp(-Div(P||Q))$ and Student-t motivated similarity $(1+Div(P||Q))^{-1}$ for distribution $P$ and $Q$ as shown in Eq. \ref{eq:gaussian} and Eq. \ref{eq:student-t}. 
	If the variance of $P$ and $Q$ are the same, the Student-t kernel based loss returns a greater loss than the Gaussian kernel based loss, when the disentanglement is not enough, as shown in Figure \ref{fig:divergence_change} and Proposition \ref{eq:prop_same_var}. Therefore, Student-t kernel contrastive loss may enforce more rigorous disentanglement. In the following propositions, $Div$ denotes the KL divergence.
	\begin{proposition}
		Assume that univariate random variables $z_{1}$ and $z_{2}$ follow Gaussian distribution $\calN(\mu_{1},\sigma^{2})$, and $\calN(\mu_{2},\sigma^{2})$, respectively. 
		Then $(1+Div(p(z_{1})||p(z_{2})))^{-1} \geq \exp(-Div(p(z_{1})||p(z_{2})))$.
		\label{eq:prop_same_var}
	\end{proposition}
	We can derive similar results when the mean of $P$ and $Q$ are the same, as shown in Proposition \ref{eq:prop_same_mean}. These rigorous disentanglement properties of Student-t based methods are effective in the general case, but they may overfit when the amount of data is restricted.
	
	Besides, this phenomenon can also occur when we handle noisy datasets. A noisy dataset, which contains corrupted labels as well as some outliers, might have a relatively large variance \cite{an2015variational}. Proposition \ref{eq:prop_same_mean} ii) denotes that the Student-$t$ based methods enforce rigorous disentangling even though the variance of $Q$ is large enough. Therefore, this strict enforcement of disentanglement can also lead to overfitting in a noisy dataset, too. We provide an empirical validation on noisy settings in Section \ref{s:results_fair}.
	\begin{proposition} 
		Assume that univariate random variables $z_{1}$ and $z_{2}$ follow Gaussian distribution $\calN(\mu_{1},\sigma_{1}^{2})$, and $\calN(\mu_{2},\sigma_{2}^{2})$, respectively. Then,
		i) the global minimum of $(1+Div(p(z_{1})||p(z_{2})))^{-1} - \exp(-Div(p(z_{1})||p(z_{2})))$ is zero if $\mu_{1}=\mu_{2}$, and ii) $\lim_{\sigma_{2} \to \infty}(1+Div(p(z_{1})||p(z_{2})))^{-1} - \exp(-Div(p(z_{1})||p(z_{2})))>0$.
		\label{eq:prop_same_mean}
	\end{proposition}
	
	\section{Experiments}
	\label{s:results}
	To validate FarconVAE, we define three research questions in terms of disentangled fairness, debiasing pretrained large-scale models, and domain generalization. We answer each question in Sections \ref{s:results_fair}, \ref{s:results_sentence}, and  \ref{s:results_domain}, respectively\footnote{We release the code at: \href{https://github.com/changdaeoh/FarconVAE}{https://github.com/changdaeoh/FarconVAE}}. See Appendix \ref{a:results_setup} for detail setup.
	\\
	RQ1) \textit{Disentangled Fairness}: Do latent representations $z_{x}$ contain non-sensitive important information only, excluding sensitive information, while maintaining the predictive performance? \\ 
	RQ2) \textit{Pretrained Models Debiasing}: Can FarconVAE be utilized for debiasing a pretrained large-scale model such as BERT? \\
	RQ3) \textit{Domain Generalization}: Does FarconVAE alleviate the spurious correlation as well as unfairness?

	\subsection{Fair Classification}
	\label{s:results_fair}
	\begin{figure}[h!]
		\begin{subfigure}{0.235\textwidth}
			\includegraphics[width=0.95\textwidth]{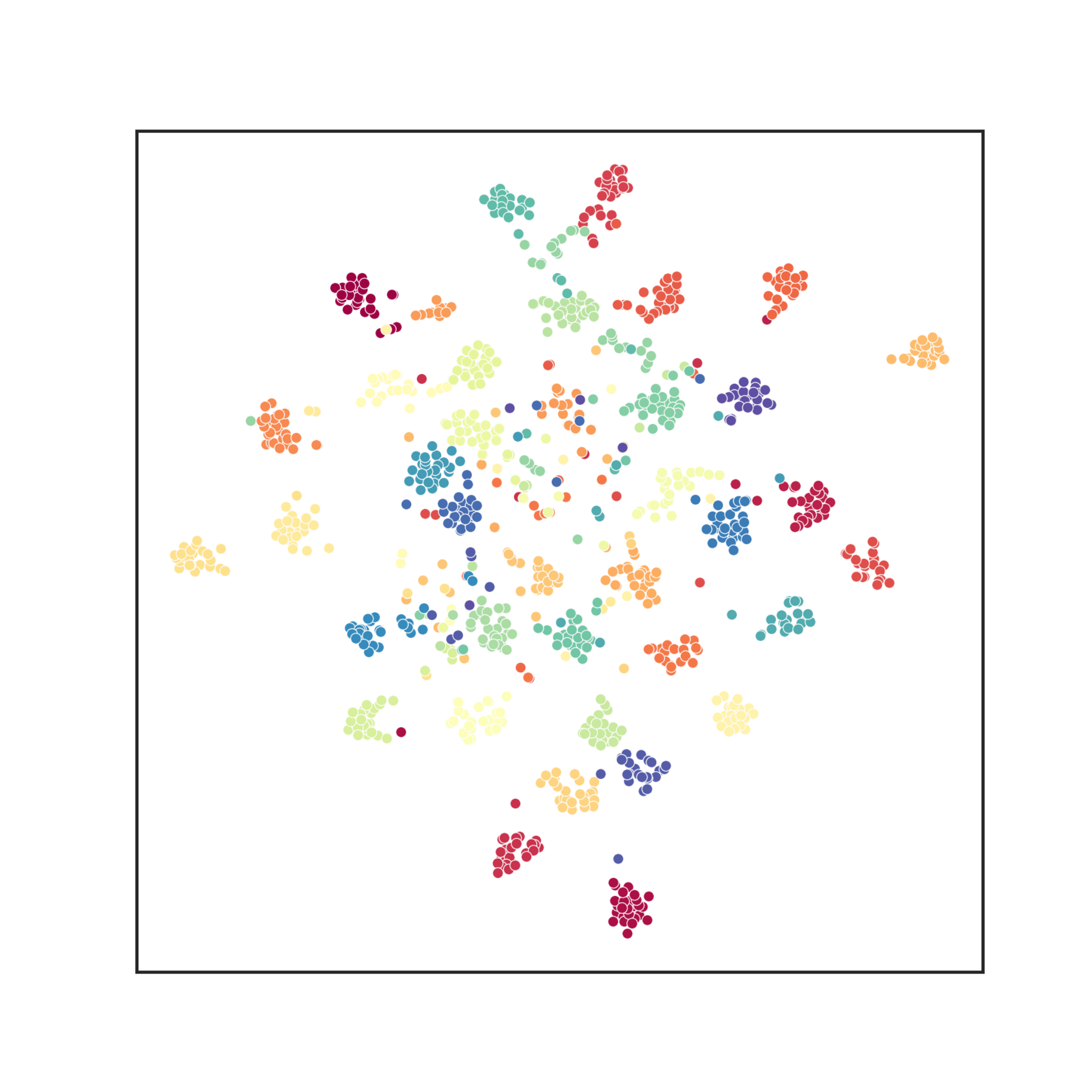}
			\caption{$z_{x}$ in ODFR (color: $y$)}
		\end{subfigure}
		\begin{subfigure}{0.235\textwidth}
			\includegraphics[width=0.95\textwidth]{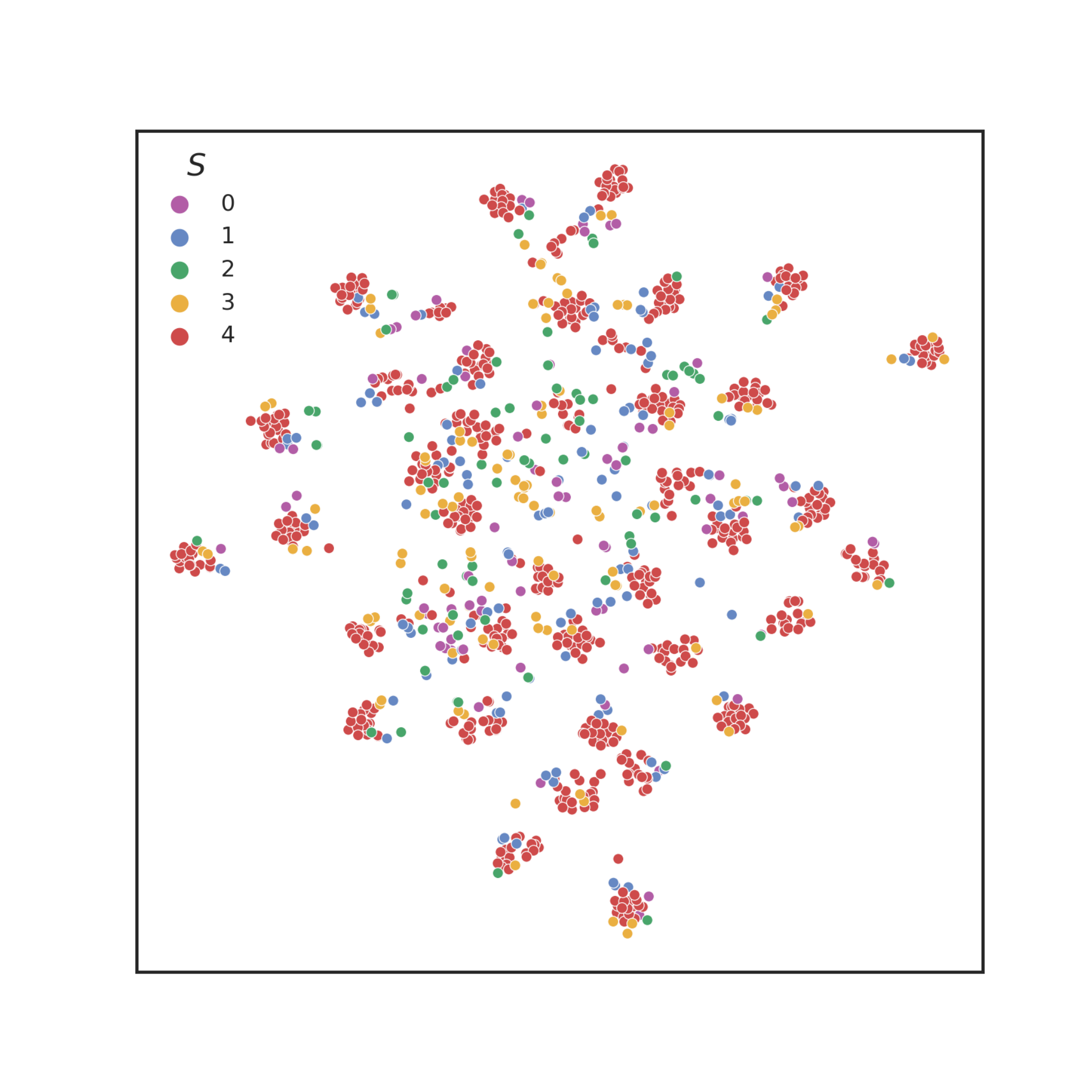}
			\caption{$z_{x}$ in ODFR (color: $s$)}
		\end{subfigure} \hfill
		\begin{subfigure}{0.235\textwidth}
			\includegraphics[width=0.95\textwidth]{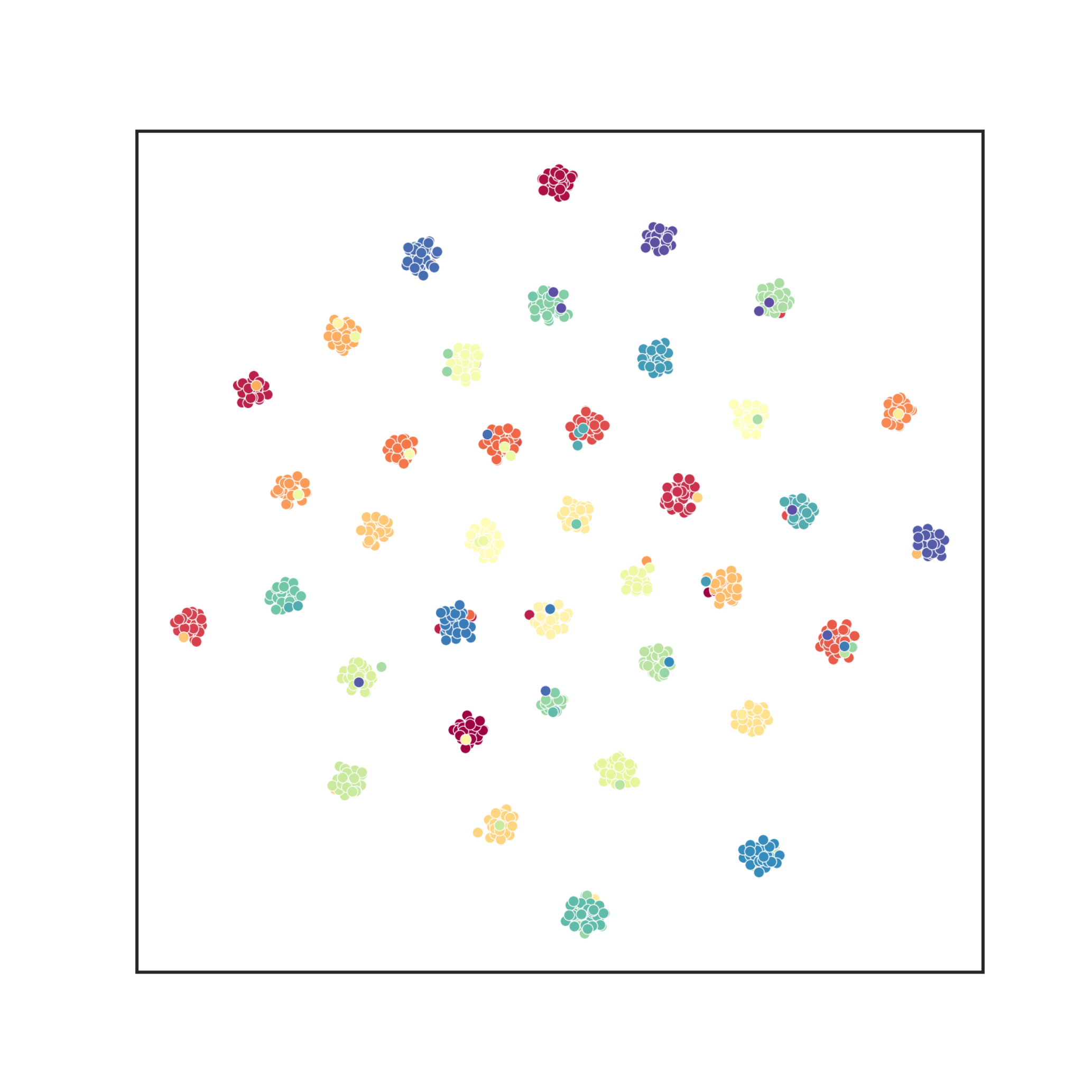}
			\caption{$z_{x}$ in FarconVAE (color: $y$)}
		\end{subfigure}
		\begin{subfigure}{0.235\textwidth}
			\includegraphics[width=0.95\textwidth]{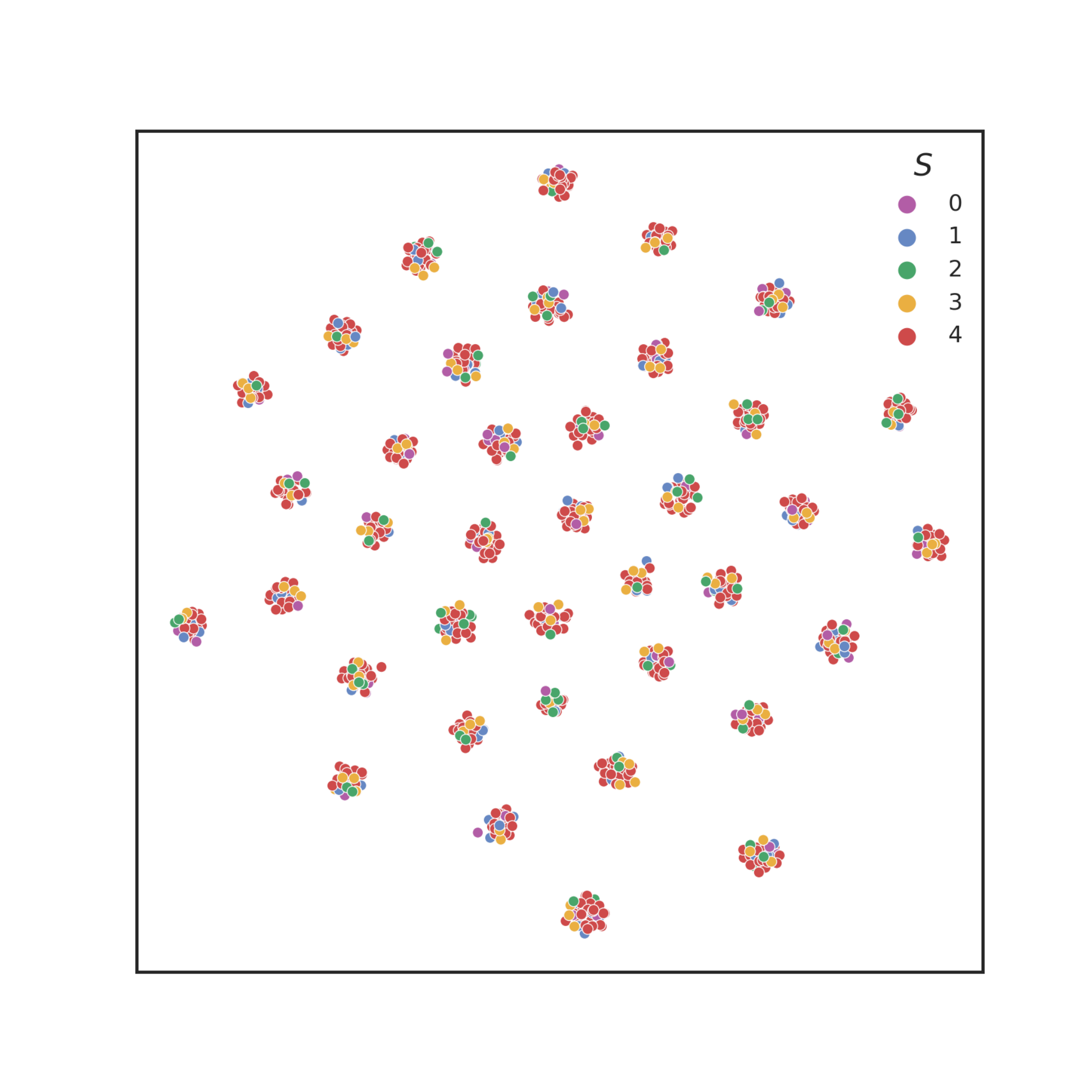}
			\caption{$z_{x}$ in FarconVAE (color: $s$)}
		\end{subfigure}
		\caption{t-SNE visualization of representation learned from ODFR and FarconVAE on YaleB. From (a) and (c), the representation of FarconVAE is more distinctive for label $y$. (b) and (d) show similar patterns, but the representation of FarconVAE shows less distinctive with respect to sensitive information $s$, and it denotes the fair representation.}
		\label{fig:tsne_fair}
	\end{figure}
	
	Table \ref{tab:fair_classification} denotes the performance on Adult, German, and Extended YaleB \cite{georghiades2001few} datasets. Adult and German are tabular datasets \cite{asuncion2007uci}, and their targets are binary about income and credit risk, respectively. Gender is a sensitive attribute for both datasets. The YaleB dataset is a visual dataset, and the target task is to classify the facial identity as irrelevant to the light condition that is regarded as a sensitive attribute. FarconVAE-G and FarconVAE-t denote the FarconVAE with Gaussian and Student-t kernel contrastive loss. We report the mean and standard deviation of 10 runs. To measure $s$ accuracy, we first train the FarconVAE and encode the entire dataset, and then train a linear classifier for $s$ on it. The representation that has similar $s$ accuracy with Random-Guess can be interpreted as fair.
	
	In terms of $y$ and $s$ accuracy, our FarconVAEs significantly improve the performance over baselines. They show the highest $y$ accuracy while $s$ accuracy is the closest with the Random-Guessing. It denotes that $z_{x}$ extracted by our model contains the non-sensitive core information while removing the sensitive information. Note that FarconVAE-t has poor performance on $s$ accuracy in German dataset, which is relatively small. Thus, the Student-t kernel based contrastive learning may overfit, while its Gaussian kernel based counterpart still works well. Besides, we visualize the learned representation of FarconVAE and ODFR. Figure \ref{fig:tsne_fair} denotes that $z_{x}$ of FarconVAE has important information to classify the target label $y$, while successfully removing the sensitive information about $s$.
	
\begin{figure}
		\centering
		\begin{tabular}{cc}
			German & YaleB  \\
			\adjustbox{valign=m,vspace=0.3pt}{\includegraphics[width=.45\linewidth]{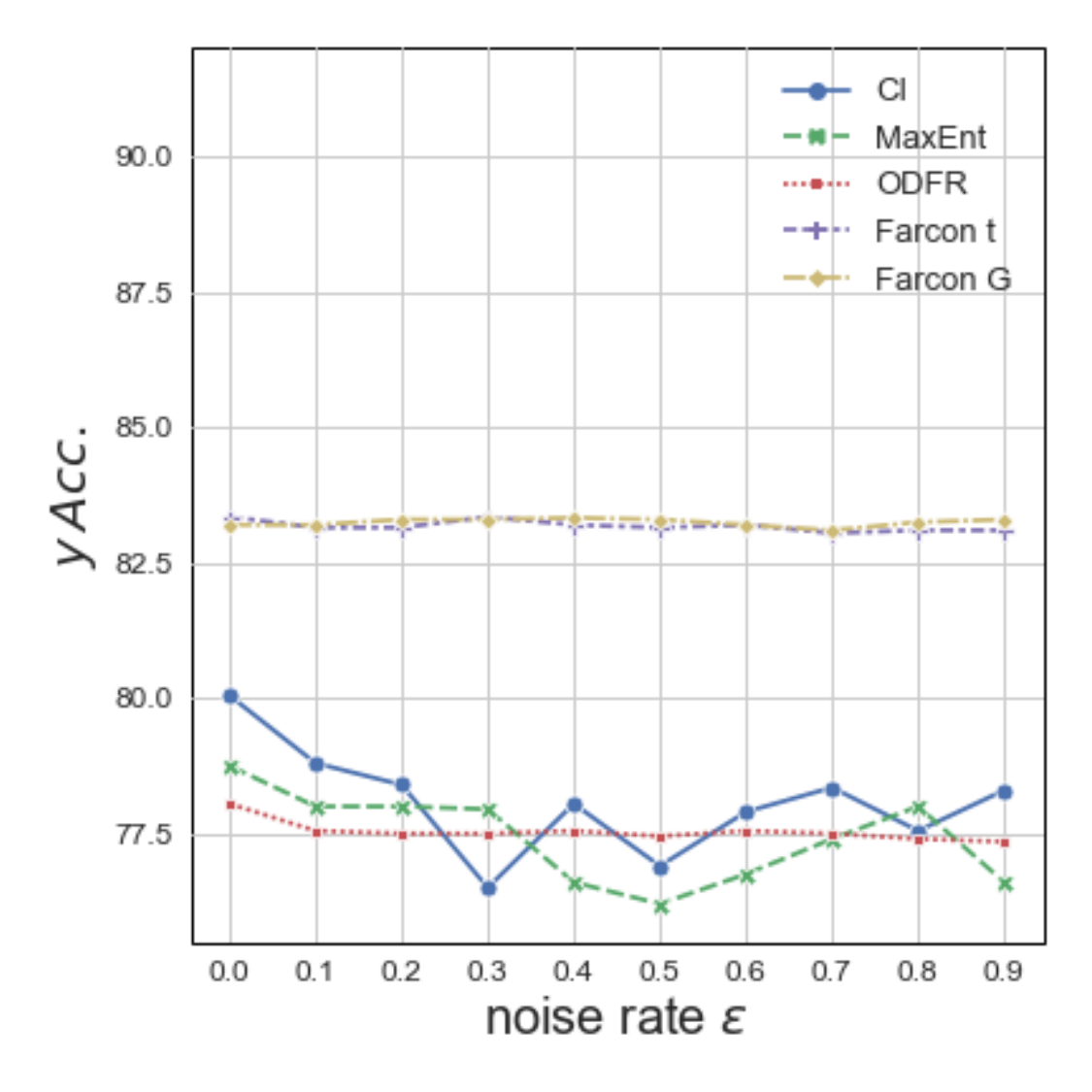}} &
			\adjustbox{valign=m,vspace=0.3pt}{\includegraphics[width=.45\linewidth]{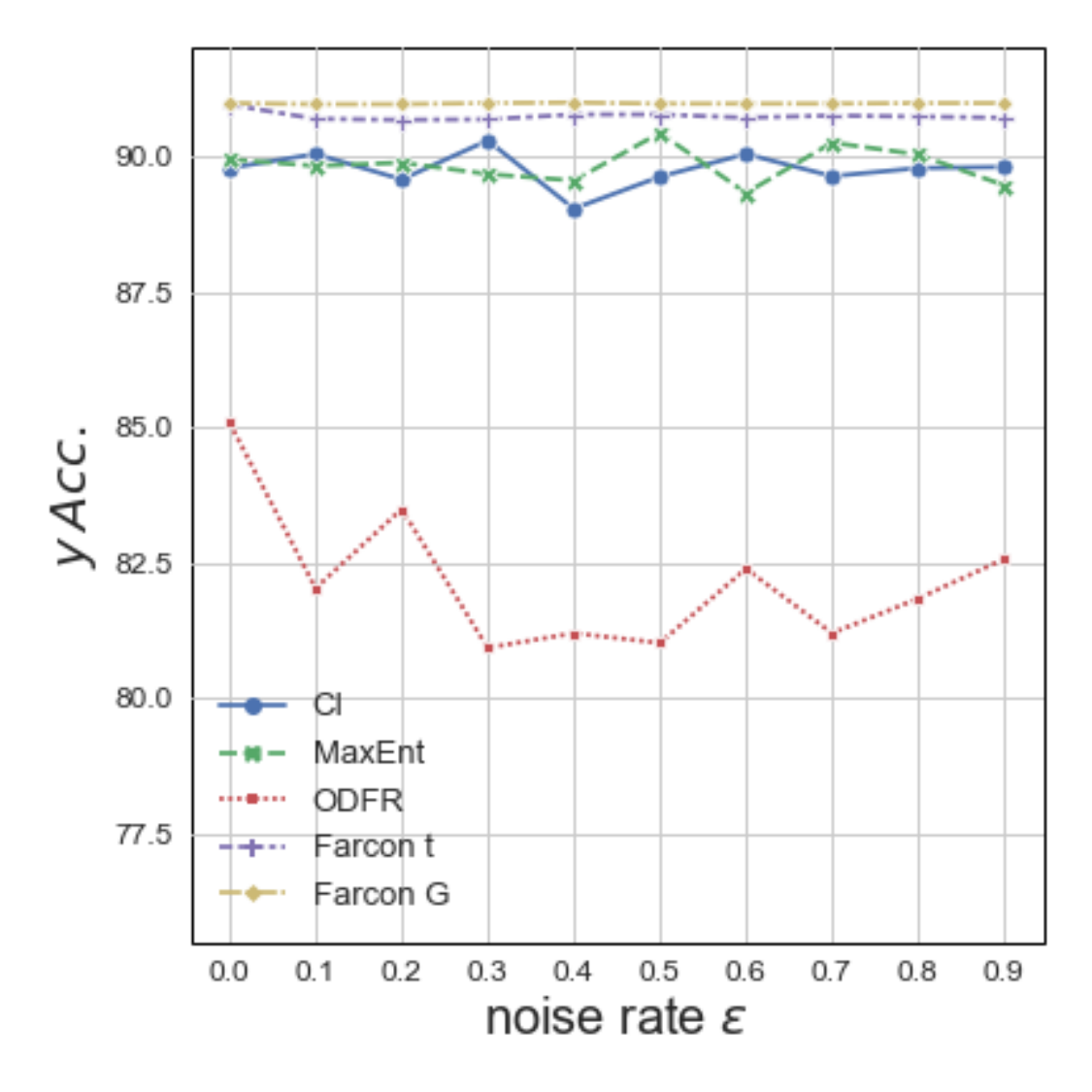}} \\
			\adjustbox{valign=m,vspace=0.3pt}{\includegraphics[width=.45\linewidth]{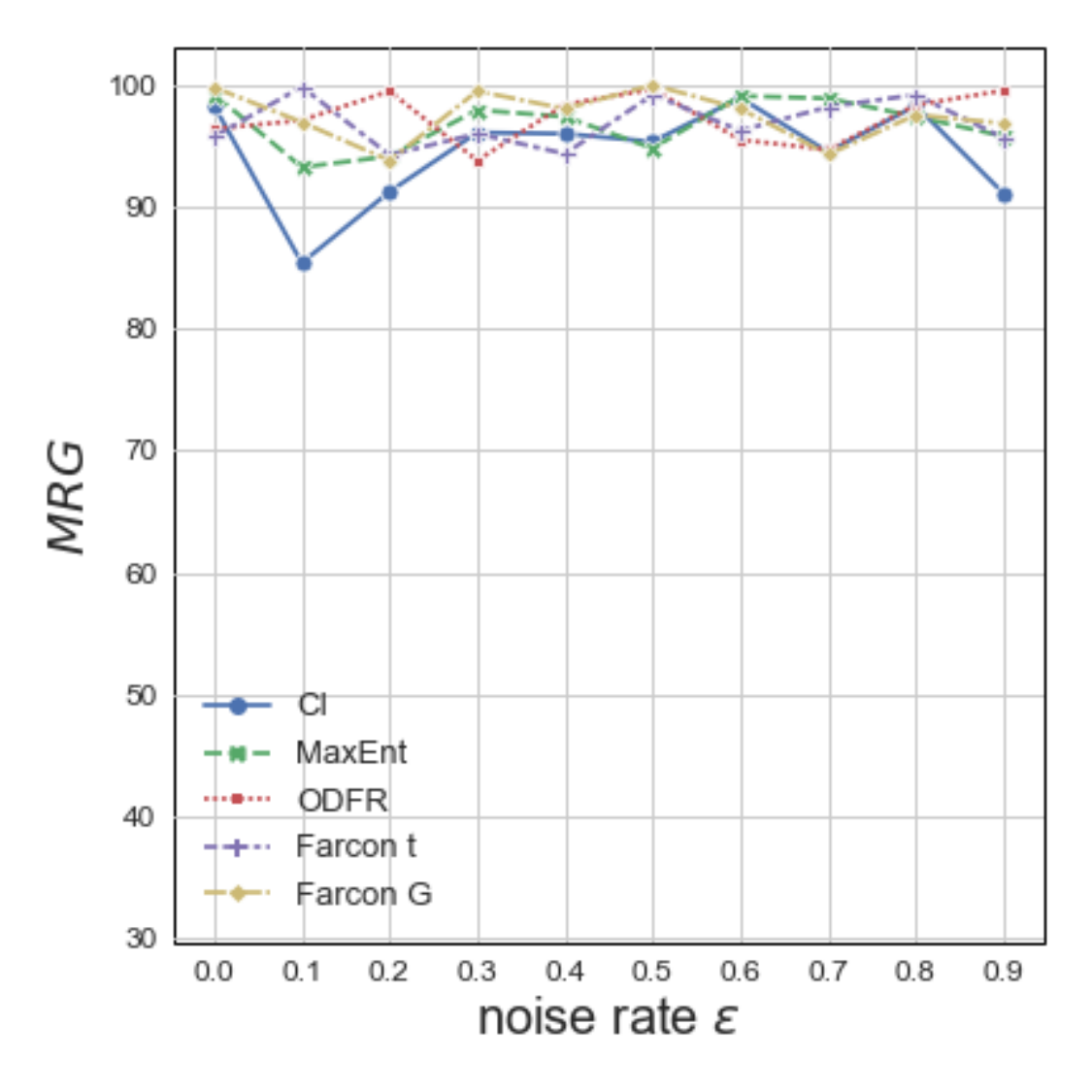}} & 
			\adjustbox{valign=m,vspace=0.3pt}{\includegraphics[width=.45\linewidth]{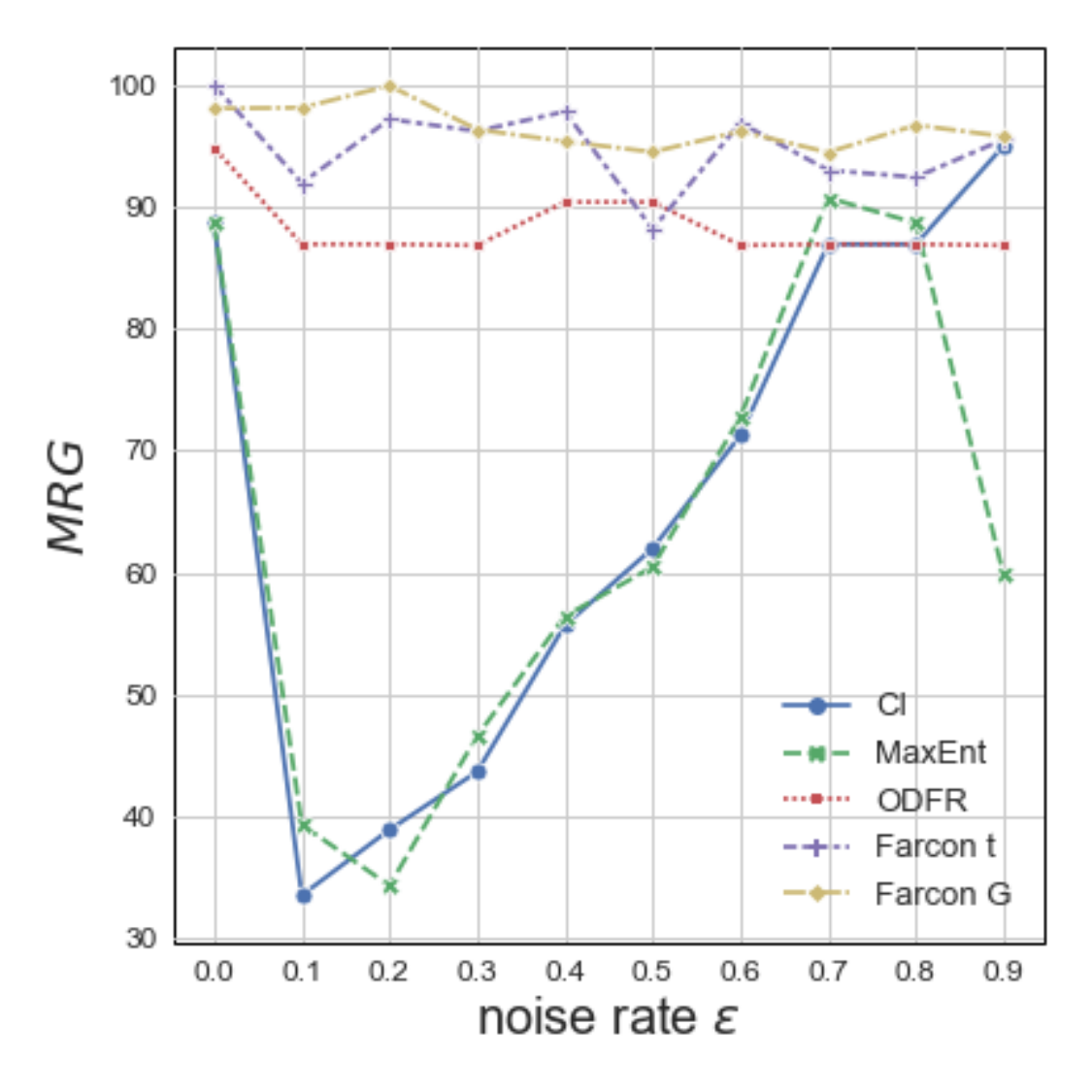}} \\
		\end{tabular}
		\caption{Performance change over noise rate $\epsilon$. Matching with Random Guess (MRG) evaluates the similarity between Random-Guess performance and model performance on $s$ accuracy, i.e., $100\%-|s \text{ } accuracy_{Random-Guess}-s \text{ } accuracy_{Model}|$. FarconVAE, especially FarconVAE-G shows a stable performance even though the noise rate $\epsilon$ is large.}
		\label{fig:noisy_setting}
	\end{figure}

	To validate the robustness of our model, we perform the evaluation on noisy data settings. 
	We construct a training set by corrupting the original sensitive attribute $s$ with another one proportional to rate $\epsilon$. Figure \ref{fig:noisy_setting} represents the performance according to the noise rate $\epsilon$.  FarconVAE-t and FarconVAE-G maintain their performance even though the $\epsilon$ increases, while other models show performance degradation. When we compare FarconVAE-t and FarconVAE-G, FarconVAE-G shows better performance than FarconVAE-t. It corresponds with our theoretical analysis, as shown in Section \ref{s:methodology_analysis}.
	
	\begin{table*}[t]
		\centering
		\caption{Debiasing performance on BERT. We use Student-t kernel for FarconVAE. $^{*}$ denotes the reported performance in \cite{cheng2020fairfil}.}
		\begin{tabular}{l|c|c|c|c|c|c|c|c|c|c|}
			\toprule
			& \multicolumn{5}{c|}{Pretrained BERT} & \multicolumn{5}{c|}{BERT post SST-2} \\
			\midrule
			&  {Origin*} &  {Sent-D*} &  {$\text{FairF}^{-}$ } &  {FairF} & {FarconVAE} &
			{Origin*} &  {Sent-D*} &  {$\text{FairF}^{-}$ } & {FairF} & {FarconVAE}  \\
			\midrule
			Names, Career/Family & 0.477 & 0.096 &  \underline{0.067} & 0.114     &  \textbf{0.003}
			& \textbf{0.036} & 0.109 & 0.142 & \underline{0.042}  & 0.101 \\
			Terms, Career/Family & 0.108 & 0.437 &   \underline{0.052} &  \textbf{0.051}  & \underline{0.052}
			& \textbf{0.010}  & \underline{0.057} & 0.136 & 0.089 & 0.097 \\
			Terms, Math/Arts & 0.253 & 0.194 &  0.154     &    \underline{0.097}  & \textbf{0.047}
			& \underline{0.219} & 0.221 & 0.423 & 0.729 & \textbf{0.034} \\
			Names, Math/Arts & 0.254 & 0.194 &  \textbf{0.015}     & 0.084      & \underline{0.056}
			& 1.153 & 0.755 & 0.769 & \underline{0.735} & \textbf{0.323} \\
			Terms, Science/Arts & 0.399 & \underline{0.075} &   0.227    &  \textbf{0.045} & 0.179
			& 0.103 & \underline{0.081} & 0.118  & 0.217 & \textbf{0.016} \\
			Names, Science/Arts & 0.636 & 0.540  & \underline{0.424}    &  0.518     & \textbf{0.005}
			& 0.222 & \underline{0.047} & 0.172 & 0.077 & \textbf{0.045} \\
			Avg. Abs. Effect Size & 0.354 & 0.256  &  0.154 & \underline{0.151} & \textbf{0.057}
			& 0.291 & 0.212 &  0.293  & \underline{0.163}  & \textbf{0.103} \\
			Classification Acc.   &    -   &    -   &    -   &    -   &  -
			&    \textbf{92.7}   &  \underline{89.1}     &    87.557   & 87.193 & 85.468 \\
			\bottomrule
		\end{tabular}
		\label{tab:pretrained_bert}%
	\end{table*}
	\subsection{Debiased Sentence Representation}
	\label{s:results_sentence}
	We evaluate our model on a text dataset as well as a tabular and image dataset. It is known that the traditional word embedding or pretrained language model such as BERT \cite{kenton2019bert} has a harmful bias. We validate whether our model, FarconVAE can debias the sentence representation, pretrained from BERT or not. Following the same experiment and evaluation setting with \cite{cheng2020fairfil}, when the input sentence contains a sensitive word, we replace it with the word with the opposite semantic meaning from the pre-defined sensitive word dictionary to construct a contrastive pair. We use the absolute SEAT effect size \cite{liang2020towards} as a measure of bias. Table \ref{tab:pretrained_bert} denotes the results for our model and baseline models including original BERT, Sent-D \cite{liang2020towards}, and FairFil \cite{cheng2020fairfil}. Like FairFil, our FarconVAE acts like a small filter that takes the BERT's sentence representation as input and outputs the debiased representation. When we adopt FarconVAE on BERT, the degree of bias reduces from 0.354 to 0.057. Another column, "BERT post SST-2", denotes the fine-tuning performance with debiased representation. FarconVAE makes relatively poor classification, but it alleviates the bias largely from 0.291 to 0.103. The results show that FarconVAE can better remove bias in a large-scale language model than existing methods.
	\subsection{Domain Generalization}
	\label{s:results_domain}
	Domain Generalization requires the invariant representations \cite{arjovsky2019invariant} or robust representation \cite{sagawa2019distributionally}, and it is commonly known that IRM and Group-DRO improve the domain generalization by alleviating the spurious correlation. However, there is no guarantee that the representations learned by IRM and Group-DRO are free from spurious correlation.
	We observe that the IRM and Group-DRO improve the predictiveness of representation for $y$ over Empirical Risk Minimization (ERM), but the representations from IRM and Group-DRO still have sensitive information largely. Table \ref{table:domain_generalization} denotes the performance of $y$ accuracy and $s$ accuracy on cMNIST \cite{arjovsky2019invariant} and Waterbirds\footnote{For a consistent evaluation, we report average accuracy for $y$ and $s$. It is different from the weighted average accuracy (for y) reported in \cite{sagawa2019distributionally} for Waterbirds dataset.} \cite{wah2011caltech} those are intentionally constructed to have a spurious correlation between $y$ and $s$. On Waterbirds, we additionally report \textit{the worst y acc.} which measures the worst accuracy among four groups distinguished by the $(s, y)$ combinations. We applied FarconVAE on top of the feature extractor trained with IRM (for cMNIST) or Group-DRO (for Waterbirds) method. Our method successfully disentangles and removes $s$ information from the learned representation by IRM or Group-DRO, so the FarconVAE is free from spurious correlation and significantly improves the $y$ accuracy.
	
	\begin{table}
		\setlength\tabcolsep{1pt}
		\smallskip
		\caption{Domain generalization performance on cMNIST and Waterbirds. Mean and standard deviation of five runs.} 
		\begin{tabular}{l|cc|ccc}
			\toprule
			& \multicolumn{2}{c}{cMNIST} & \multicolumn{3}{c}{Waterbirds}\\ \midrule
			Model & $y$ acc. & $s$ acc. & $y$ acc. & worst $y$ acc. & $s$ acc. \\ \midrule
			ERM & 16.91$\pm{0.57}$ & \underline{97.65$\pm{0.21}$} & 78.47$\pm{0.57}$& 31.87$\pm{1.20}$ & \underline{86.07$\pm{0.69}$} \\
			IRM & \underline{66.27$\pm{1.42}$} & 100.00$\pm{0.00}$ & - & - & - \\
			gDRO & - & - & \underline{90.21$\pm{0.57}$} & \underline{85.98$\pm{1.16}$} &86.60$\pm{0.91}$ \\
			Ours & \textbf{70.43$\pm{0.61}$} & {\textbf{35.10$\pm{3.35}$}}&{\textbf{95.03$\pm{0.99}$}}& \textbf{91.71$\pm{1.48}$} &\textbf{51.71$\pm{0.38}$}\\
			\bottomrule
		\end{tabular}
		\label{table:domain_generalization}
	\end{table}
	\begin{figure}
		\label{table:dg_tsne}
		\centering
		\begin{tabular}{cccc}
			& $s$ & $y$ & Predicted $y$ \\
			ERM  & \adjustbox{valign=m,vspace=0.3pt}{\includegraphics[width=.25\linewidth]{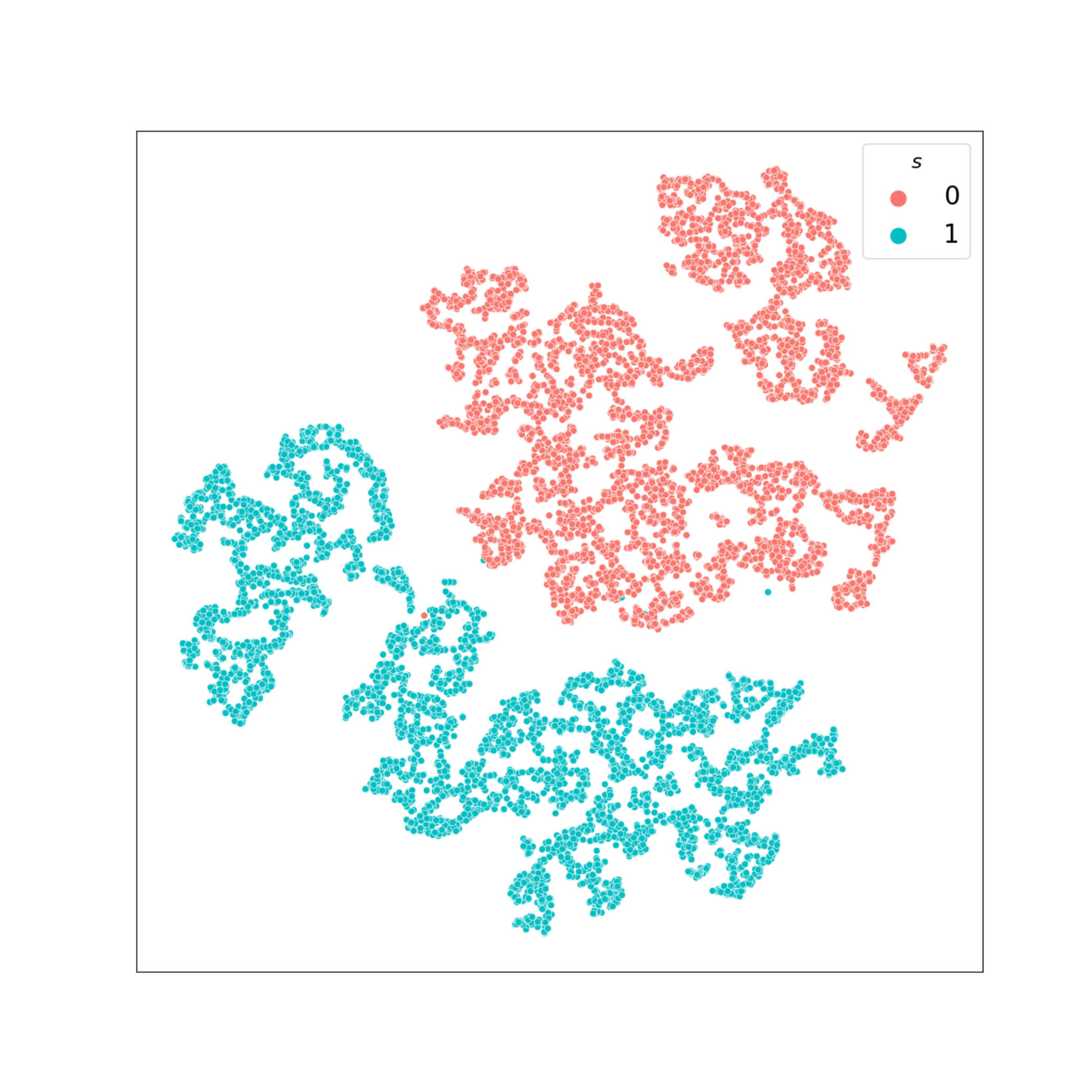}} & \adjustbox{valign=m,vspace=0.3pt}{\includegraphics[width=.25\linewidth]{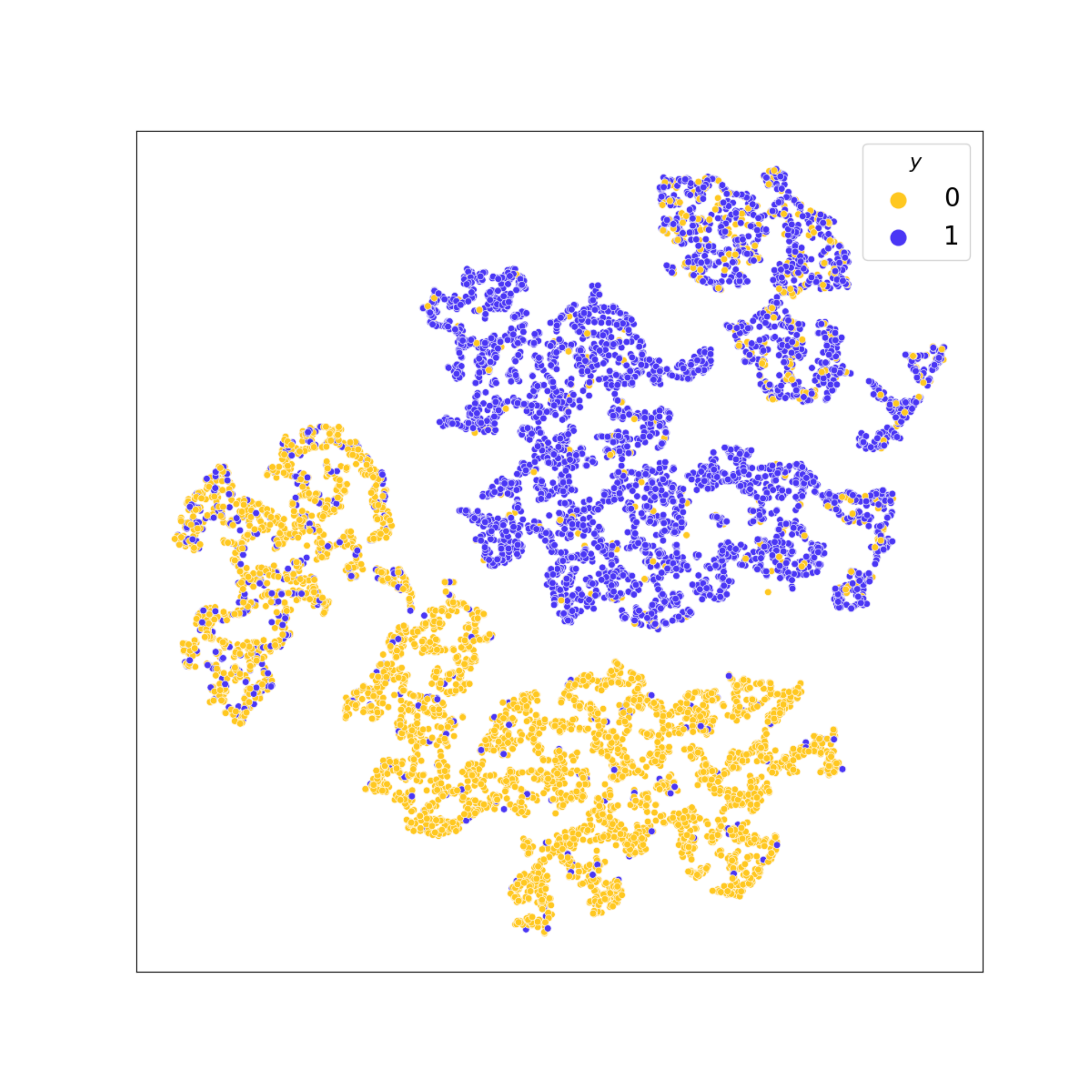}} & \adjustbox{valign=m,vspace=0.3pt}{\includegraphics[width=.25\linewidth]{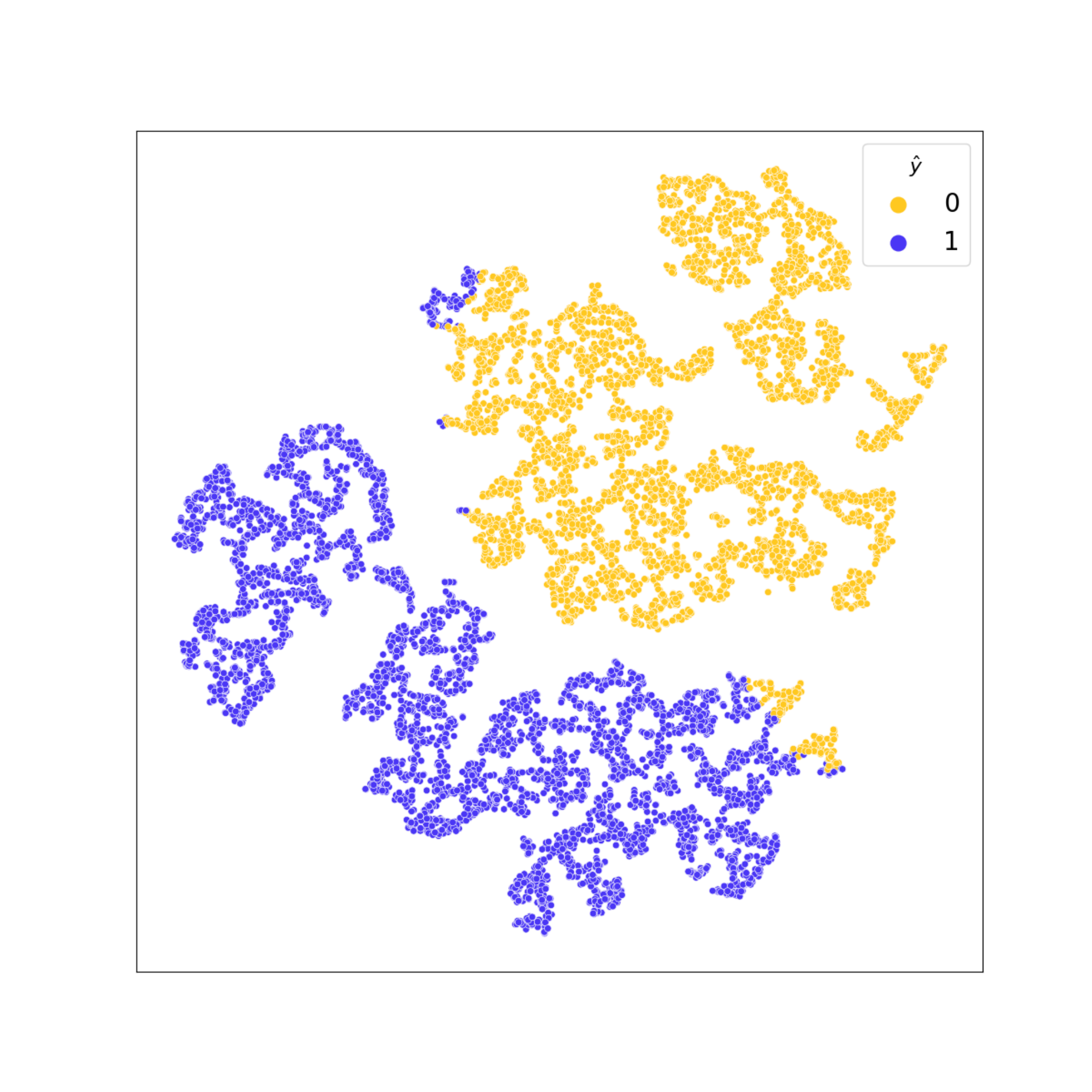}} \\
			IRM  & \adjustbox{valign=m,vspace=0.3pt}{\includegraphics[width=.25\linewidth]{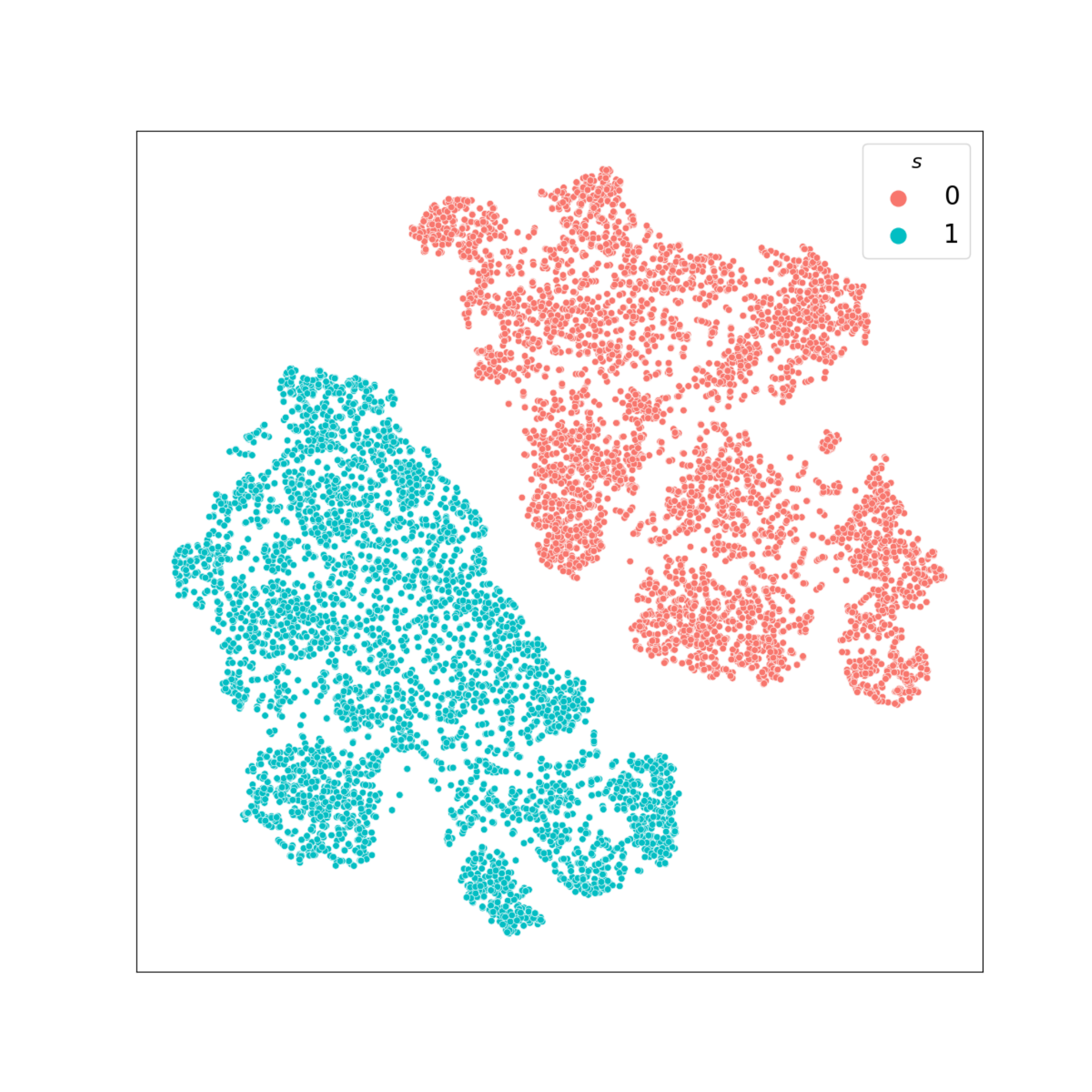}} & \adjustbox{valign=m,vspace=0.3pt}{\includegraphics[width=.25\linewidth]{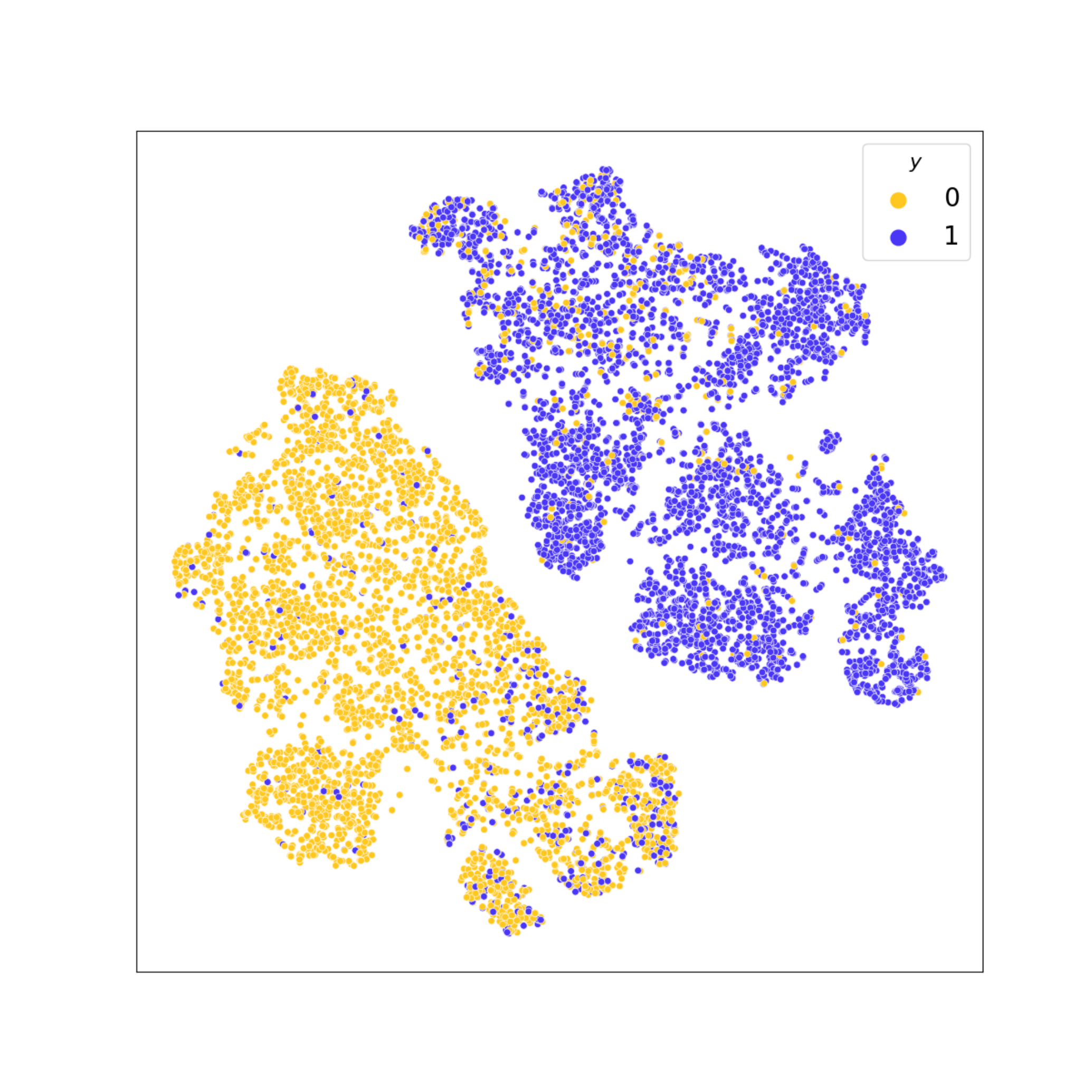}} & \adjustbox{valign=m,vspace=0.3pt}{\includegraphics[width=.25\linewidth]{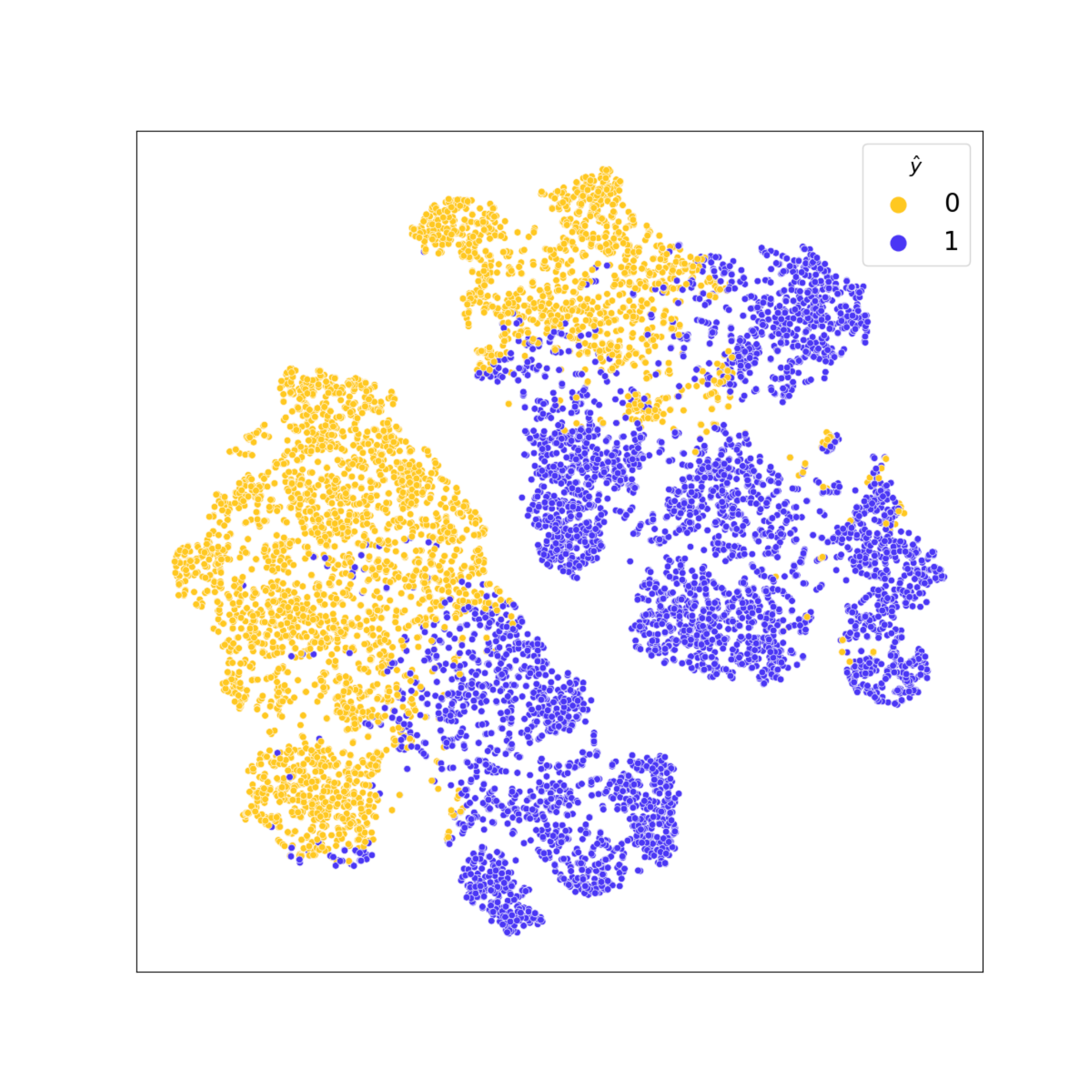}}  \\
			Ours  & \adjustbox{valign=m,vspace=0.3pt}{\includegraphics[width=.25\linewidth]{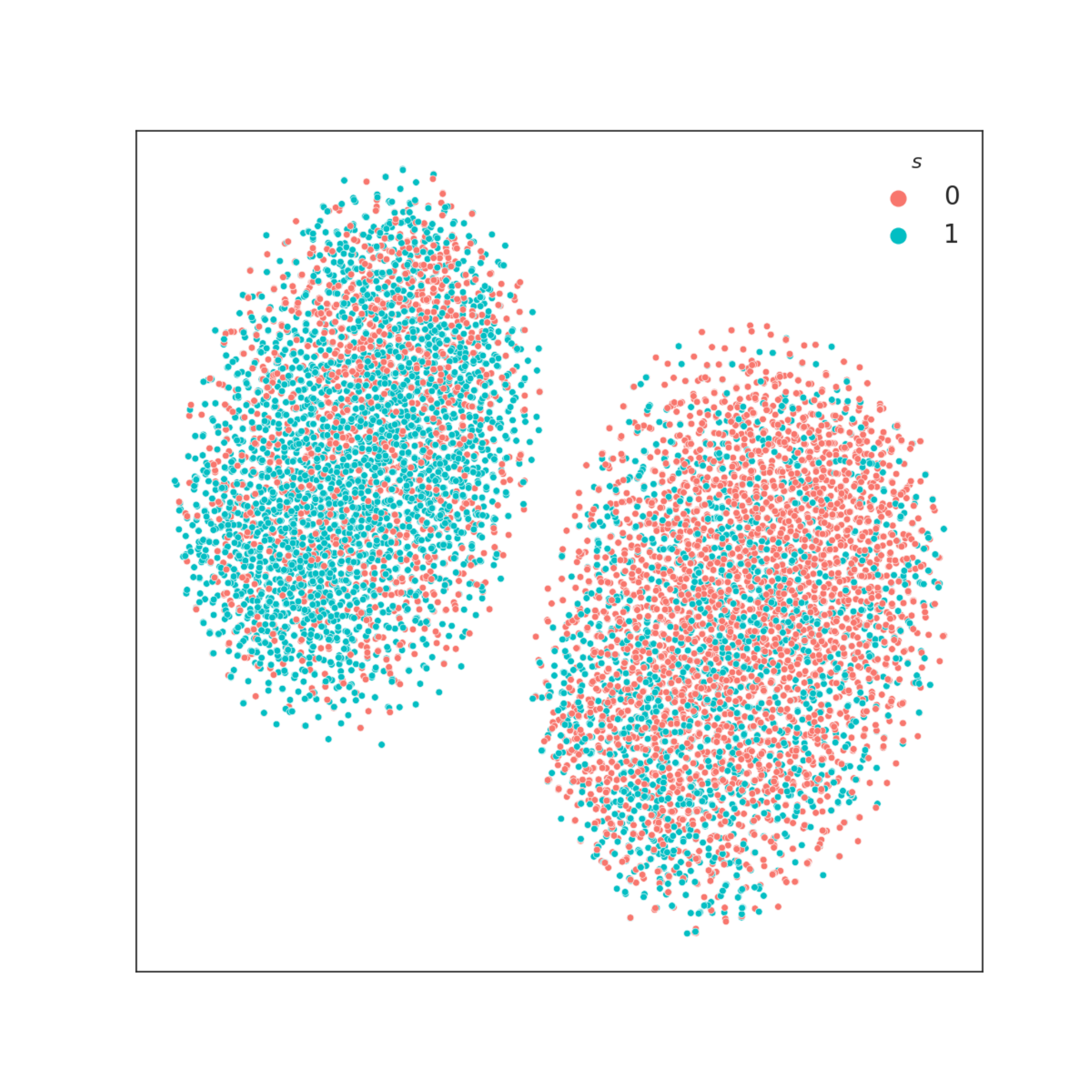}} & \adjustbox{valign=m,vspace=0.3pt}{\includegraphics[width=.25\linewidth]{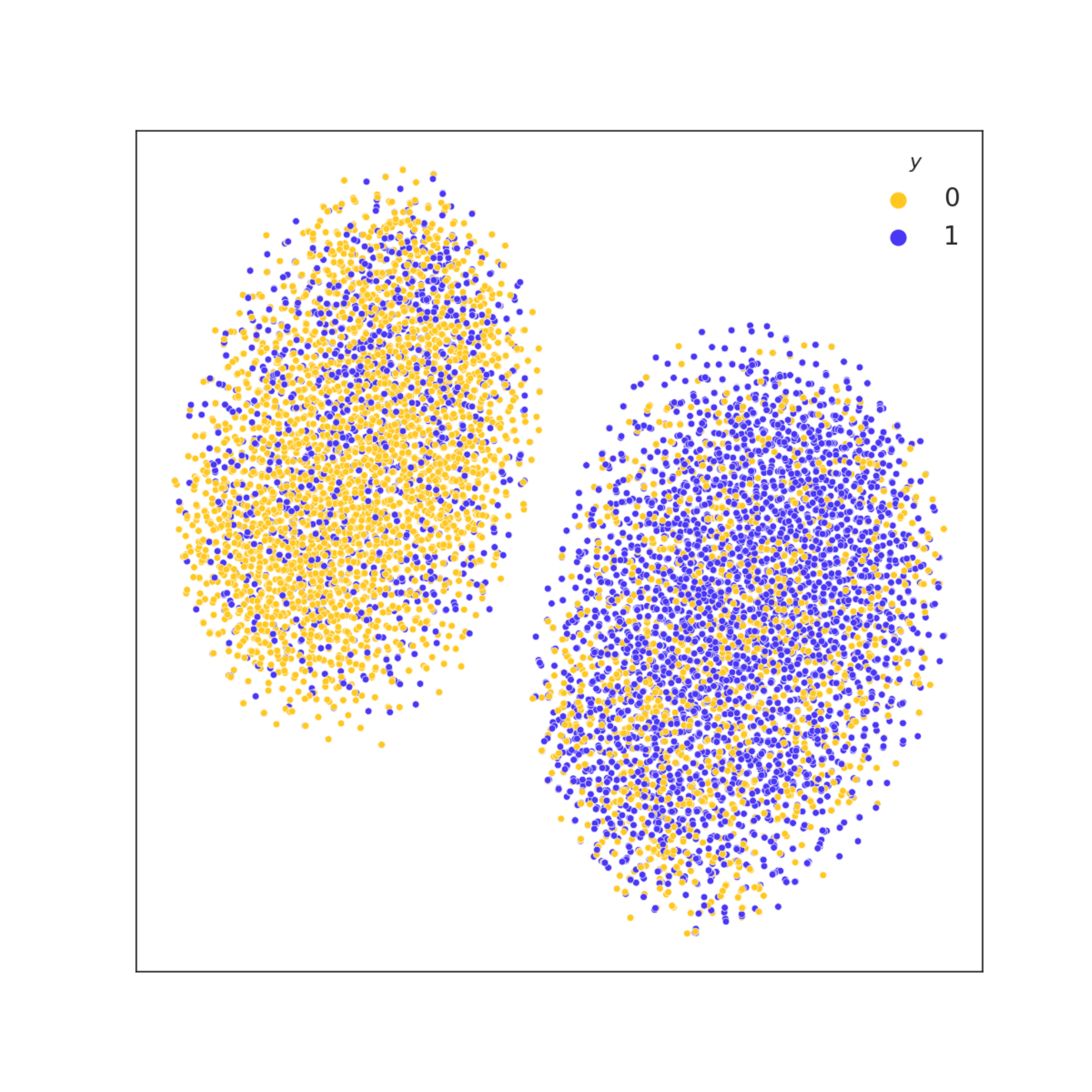}} & \adjustbox{valign=m,vspace=0.3pt}{\includegraphics[width=.25\linewidth]{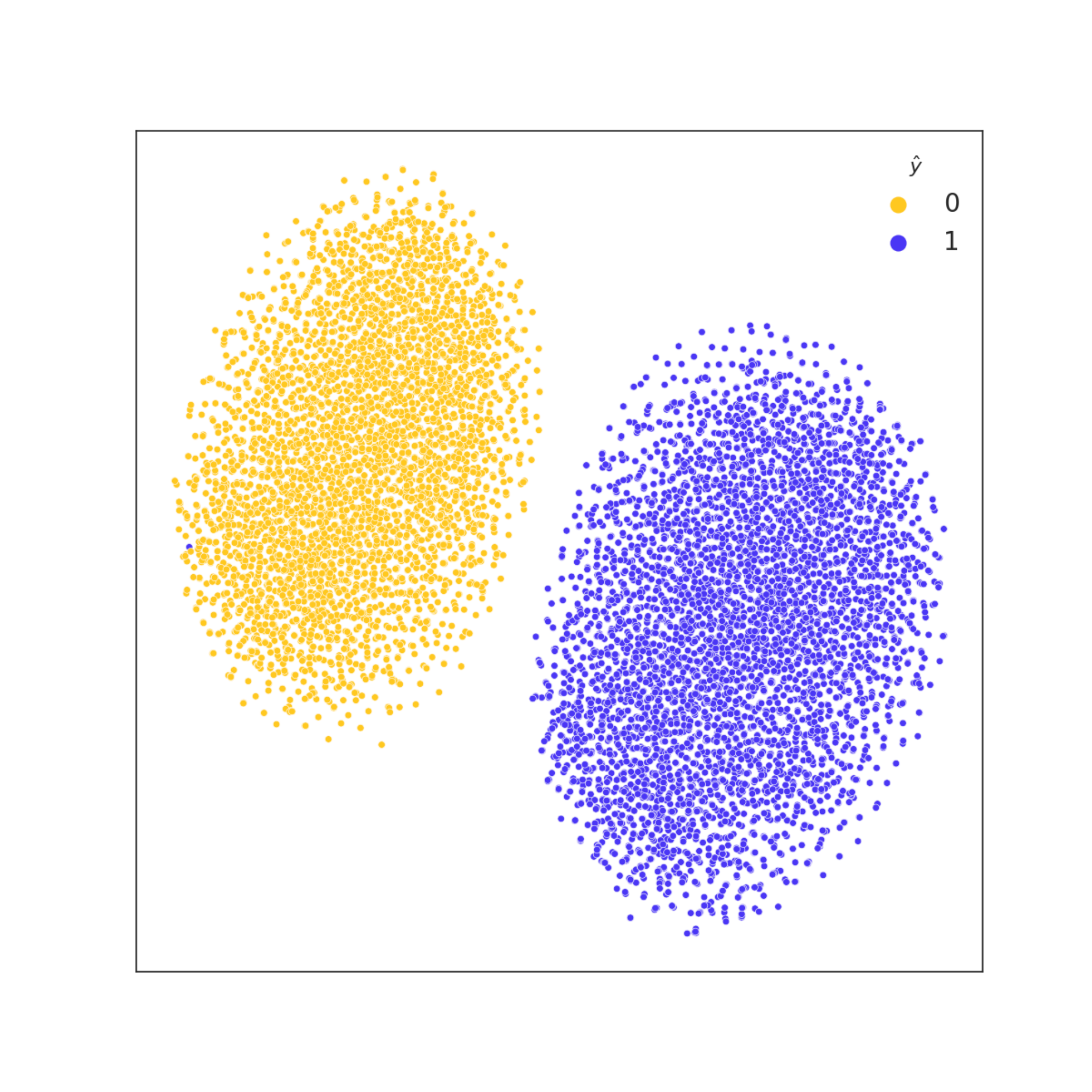}} 
		\end{tabular}
		\caption{t-SNE visualization for representation of the penultimate layer from ERM, IRM, and FarconVAE. As shown in the first column, the representations by ERM and IRM have sensitive information, while that from FarconVAE has significantly less sensitive information. Because the representation from ERM and IRM has spurious correlation, they have difficulty in predicting $y$ in test time when the correlation is reversed, while FarconVAE shows stable prediction.} 
	\end{figure}
	Figure 6 denotes the visualization of learned embedding from ERM, IRM, and FarconVAE on cMNIST. As shown in the first column of Figure 6, the learned representation from ERM and IRM are easily divided by sensitive attribute $s$. 
	However, the representation learned from FarconVAE is not easily divided by sensitive attributes. From the second and third columns, we can check that FarconVAE predicts $y$ well, while ERM and IRM predict opposite $y$ entirely or partially, respectively. To the best of our knowledge, this is the first work that provides general disentangling methods that can be utilized in both domain generalization and fairness tasks.
	\section{Conclusion}
	\label{s:conclusion}
	Algorithmic fairness demands fair representation learning that captures the non-sensitive core information. Domain generalization tasks also have difficulty removing spurious features for domain-invariant learning.
	This paper proposes FarconVAE, a new contrast-based disentangling approach that removes sensitive information or spurious correlation from representation while maintaining non-sensitive core information.
	We provide a kernel motivated distributional contrastive loss that stably induces disentangled invariant representation and a swap-recon loss that enforces swap-consistent reconstruction for further enhancing disentanglement.
	
	FarconVAE improves the out-of-distribution generalization as well as the fairness. Besides, we observe that the representation learned by IRM or Group-DRO still has a strong spurious correlation between the targets and sensitive attributes, and FarconVAE mitigates this correlation significantly. We provide extensive empirical results about fairness, pretrained model debiasing, and domain generalization on tabular, image, and text datasets. Moreover, we derive a theoretical result for our new contrastive loss.
	
	\section*{Acknowledgements}
	This work was partly supported by Institute of Information \& Communications Technology Planning \& Evaluation (IITP) grant funded by the Korea government(MSIT) (No.2021-0-02067,Next generation AI for multi-purpose video search,50\%) and the National Research Foundation of Korea(NRF) grant funded by the Korea government(MSIT). (No. 2021R1F1A1060117, Research and Application of Artificial Intelligence Algorithm for Removing Spurious Bias,50\%)
	
	\bibliographystyle{ACM-Reference-Format}
	\bibliography{FarconVAE}
	
	\newpage
	\appendix
	
	\section{Experimental Setup}
	\label{a:results_setup}
	\subsection{Fair Classification}
	For fair classification, we consider three benchmark datasets previously used in \cite{sarhan2020fairness}. The Adult dataset contains 45,222 instances, each with 14 attributes. The target $y$ is a binary label of annual income more or less than \$50,000 and gender is sensitive attribute $s$. The German dataset has 1,000 instances, each with 20 attributes, and the target task is to classify bank account holders with good or bad credit risk. The sensitive attribute is gender again. Both of these are tabular datasets obtained from the UCI ML-repository \cite{asuncion2007uci}. The Extended YaleB \cite{georghiades2001few} is a visual dataset that contains the face images of 38 people under five different light conditions. The target task is to identify one of the 38 people for a given data instance, while the light condition is the sensitive attribute here.
	
	\subsection{Pretrained Model Debiasing}
	For the pretrained model debiasing task, we validate our method on BERT \cite{kenton2019bert}. Specifically, we attach the FarconVAE-t on top of BERT's layers, takes [CLS] token embedding for each sentence as model input $x$ and the sensitive words are regarded as $s$. Following the setup \cite{cheng2020fairfil}, we use the same corpora consisting of 183,060 sentences for training, and the sensitive attribute is mainly gender-related words.
	
	\subsection{Domain Generalization}
	For domain generalization task, we experimented with two image datasets cMNIST \cite{arjovsky2019invariant} and Waterbirds \cite{sagawa2019distributionally}. The cMNIST is a synthetic dataset for binary classification, which intentionally makes correlation between labels (digit) and sensitive attributes (color) of the train set. A model is evaluated with the test set, which has the opposite correlation to the train set. The Waterbirds dataset is constructed by combining bird photographs from the CUB dataset \cite{wah2011caltech} with backgrounds from the Places dataset \cite{zhou2017places}. The target label is the bird's breed (waterbird or landbird), and the sensitive attribute is the background (water or land). Like cMNIST, there is a spurious correlation between $y$ and $s$ in the train dataset.
	
	\section{Implementation Details}
	\label{a:implementation}
	\subsection{Conditional Information at Test Time}
	We put $y$ as FarconVAE input to increase the predictiveness of the representation. But unlike the training phase, we generally do not have access to true labels in the testing phase. So we train a separate classifier predicting $y$ from $x$ in advance or together with FarconVAE's training phase. It is also possible to use a well-fitted pretrained model on the given dataset. After we build the best classifier that maps $x$ to $y$, we use the predicted label $\hat{y}$ by the classifier for each test set instance as the input of FarconVAE. For Adult, German, Extended YaleB, and CMNIST, this classifier is a simple multi-layer perceptron (MLP). For Waterbirds dataset, the classifier is ResNet-50 \cite{he2016deep}. For debiasing task on BERT, when FarconVAE is trained on the unannotated corpus (results of the left side in Table \ref{tab:pretrained_bert}), the target label does not exist. In this case, we input the constant $y=0.5$ to FarconVAE. When FarconVAE is finetuned with BERT on the labeled corpus (results of the right side in Table \ref{tab:pretrained_bert}), we also input the constant $y=0.5$ for consistent training.
	
	\subsection{Model Configuration}
	In this subsection, we introduce the setting of model architecture, hyperparameters, and other experimental options. To learn more informative representation, we use ELBO of $\beta$-VAE formulation \cite{higgins2016beta}, which can control the intensity of KLD regularization instead of the basic ELBO in Section \ref{s:methodology_farconvae}. So, our ELBO has a parameter $\beta$. 
	\begin{align}
		{L_{ELBO}(\bm{\phi},\bm{\theta}) = L_{rec}(\bm{\phi},\bm\theta\textbackslash\theta_{y}) + L_{pred}(\bm{\phi},\theta_{y}) - \beta L_{KLD}(\bm{\phi})} \nonumber
	\end{align}
	In the sections below, $\alpha$, $\beta$, $\gamma$, $LR$, and $WD$ denote the weight of loss terms $L_{DC}$, $L_{KLD}$, $L_{SR}$, learning rate, and weight decay hyperparameter, respectively. See Github\footnote{\href{https://github.com/changdaeoh/FarconVAE}{https://github.com/changdaeoh/FarconVAE}} for details not listed here.
	
	\subsubsection{Fair Classification} On Adult and German datasets, we follow the setup in \cite{roy2019mitigating, sarhan2020fairness} for all possible configurations. So, the encoder $q_{\bm\phi}$ and decoder $p_{\bm\theta \textbackslash\theta_{y}}$ are both one hidden layer with 64 hidden units and the decoder $p_{\theta_{y}}$ (refered as target predictor in \cite{roy2019mitigating, sarhan2020fairness}) is linear logistic regression. The latent dimension of $z_{x}$ are 15 in Adult and 5 in German. For Extended YaleB dataset, we use one linear layer as encoder, decoder $p_{\bm\theta \textbackslash\theta_{y}}$ and decoder $p_{\theta_{y}}$ each contain 100 hidden units. The latent dimension is also 100. 
	
	\subsubsection{Pretrained Model Debiasing} We use one linear layer for all components of FarconVAE with 128 hidden units and 128 latent dim. For doing contrastive learning on unlabeled corpus, we flip the sensitive words of a given sentence to follow \cite{cheng2020fairfil}. If a sentence does not contain any sensitive words, a constant tensor of 0.5 which has the same shape as the embedding was inputted to FarconVAE and $L_{DC}$ and $L_{SR}$ are not used. In the fine-tuning stage, most of the sentences do not have sensitive words, so we use the mean embedding of pre-defined sensitive words list as FarconVAE input. We use ($\alpha$, $\beta$, $\gamma$, $LR$, $WD$) = (1.0, 0.2, 1.0, 5e-4, 1e-4) for FarconVAE contrastive-training, and ($\alpha$, $\beta$, $\gamma$, $LR_{BERT}$, $LR_{Farcon}$, $WD$) = (1.0, 0.2, 0.0, 2e-5, 1e-4, 1e-2) for entire fine-tuning.
	
	\subsubsection{Domain Generalization} Like above, we attach the FarconVAE-t on top of IRM or Group-DRO feature extractors (fixed) and use the representation from them as FarconVAE's input feature. Again, we use one linear layer for all components of the FarconVAE latent dim set to 100 with 75 and 100 hidden units for cMNIST and Waterbirds, respectively. For cMNIST, we use ($\alpha$, $\beta$, $\gamma$, $LR$, $WD$) = (1.0, 0.2, 0.0, 1e-3, 1e-4). For Waterbirds, we use ($\alpha$, $\beta$, $\gamma$, $LR$, $WD$) = (0.5, 0.2, 0.5, 7e-4, 1e-4) and we anneal $\beta$ from zero to 0.2 during the first 10\% epochs.
	
	\section{Ablation Study}
	\label{a:results_ablation}
\begin{figure}
		\begin{subfigure}{0.235\textwidth}
			\includegraphics[width=0.95\textwidth]{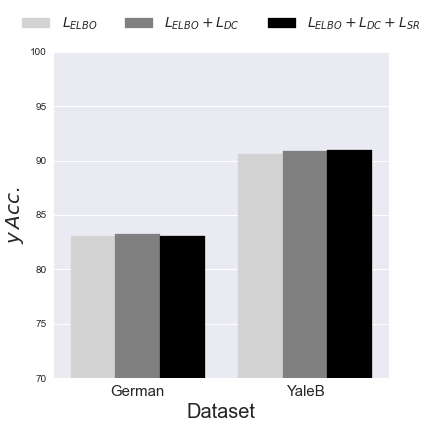}
			\caption{Ablation study for $y$ accuracy}
		\end{subfigure}
		\begin{subfigure}{0.235\textwidth}
			\includegraphics[width=0.95\textwidth]{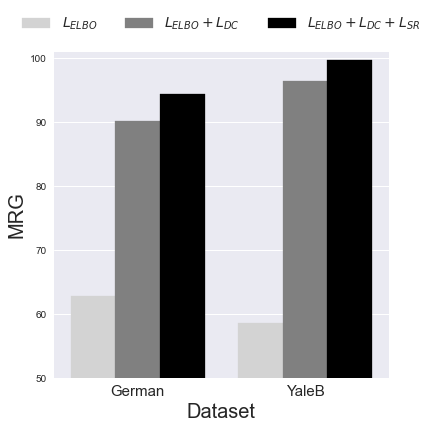}
			\caption{Ablation study for MRG}
		\end{subfigure}
		\caption{Ablation study for $\bL_{DC}$ and $\bL_{SR}$.}
		\label{fig:ablation}
	\end{figure}

	In this section, we provide the ablation study for our proposed methods. Figure \ref{fig:ablation} indicates that our proposed algorithm, FarconVAE with distributional contrastive loss is effective in disentangling the latent representation space, and as a result, it induces a fair representation. Moreover, the disentanglement is further enhanced when swap-recon ($\bL_{SR}$) is added. MRG in the right panel of Figure \ref{fig:ablation} denotes similarity between the s accuracy of model and that of random guessing, same with Figure \ref{fig:noisy_setting}.
	\section{Derivation of ELBO}
	We assume that the $z_{x}$ and $z_{s}$ are conditionally independent given $x,s,y$\footnote{To satisfy the assumption, we adopt contrastive loss as shown in Sec. 3 of main paper}, i.e., ${q_{\bm\phi}(z_x,z_s|x,s,y)=q_{\phi_{z_x}}(z_x|x,s,y)q_{\phi_{z_s}}(z_s|x,s,y)}$. Then, the evidence lower bound of the log marginal likelihood is:
	\begin{flalign}
		\small
		& \log{p_{\bm{\theta}}(x,s,y)} & \nonumber \\
		& = \log \int {\frac{q_{\bm{\phi}}(z_{x},z_{s}|x,s,y)}{q_{\bm{\phi}}(z_{x},z_{s}|x,s,y)}p_{\bm{\theta}}(x,s,y|z_{x},z_{s})p(z_{x},z_{s})dz_{s}dz_{x}} \nonumber &\\
		& \geq\int{q_{\bm{\phi}}(z_{x},z_{s}|x,s,y)
			\log \frac{p_{\bm{\theta}}(x,s,y|z_{x},z_{s})p(z_{x},z_{s})}{q_{\bm{\phi}}(z_{x},z_{s}|x,s,y)}dz_{s}dz_{x}} \nonumber &\\
		& = \E_{q_{\bm{\phi}}}[\log p_{\theta_{x}}(x|z_{x},z_{s}) + 
		\log p_{\theta_{s}}(s|z_{x},z_{s}) + \log p_{\theta_{y}}(y|z_{x})] \nonumber &\\
		& -\KL(q_{\phi_{x}}(z_{x}|x,s,y)||p(z_{x})) -\KL(q_{\phi_{s}}(z_{s}|x,s,y)||p(z_{s})) \nonumber &  \\ 
		& = \bL_{ELBO}(\bm{\phi},\bm{\theta};x,s,y) & \label{eq:FarconVAE_ELBO} \nonumber
	\end{flalign}
	\section{Proof}
	\paragraph{Proof for Proposition 1} For simplicity, we denote $t$ as a $\mu_{1}-\mu_{2}$.
	The KL divergence between two Gaussian distributions, $\calN(\mu_{1},\sigma^{2})$ and $\calN(\mu_{2},\sigma^{2})$ are $\frac{t^{2}}{2\sigma^{2}}$. \\\
	Then, $(1+Div(p(z_{1})||p(z_{2}))) - \exp(Div(p(z_{1})||p(z_{2}))) = \frac{t^{2}}{2\sigma^{2}}- \exp{(\frac{t^{2}}{2\sigma^{2}})}+1$. \\
	When $\mu_{1}=\mu_{2}$, $1+\frac{t^{2}}{2\sigma^{2}}- \exp{(\frac{t^{2}}{2\sigma^{2}})} = 0$, \\
	and $\frac{\partial(1+Div(p(z_{1})||p(z_{2}))) - \exp(Div(p(z_{1})||p(z_{2})))}{\partial t}=\frac{t-t\exp(x^{2}/(2\sigma^{2}))}{\sigma^{2}}$<0. \\
	Therefore, $(1+Div(p(z_{1})||p(z_{2})))^{-1} \geq \exp(-Div(p(z_{1})||p(z_{2})))$, and the equality holds when $t=0$, i.e., $\mu_{1}=\mu_{2}$.
	
	\paragraph{Proof for Proposition 2} For simplicity, we denote $t$ as a $\frac{\sigma_{2}}{\sigma_{1}}$. \\
	i) The KL divergence between two Gaussian distributions, $\calN(\mu,\sigma_{1}^{2})$ and $\calN(\mu,\sigma_{2}^{2})$ are $\log(t)+\frac{1}{2t^2}-0.5$. \\\
	Then, $f(t) = (1+Div(p(z_{1})||p(z_{2}))) - \exp(Div(p(z_{1})||p(z_{2}))) = 0.5 + \frac{1}{2t^{2}} - \exp(-0.5 + \frac{1}{2t^{2}})t + \log(t).$ \\
	$f(t)$ is differentiable for all $t>0$, and the only critical point of $\frac{\partial{f(t)}}{\partial{t}}=0$ is at $t=1$. The domain of $f'(t)$ is ${t \in R : t>0}$, and $f(t)$ is $-\infty$ when $t=0^{+}$ and $\infty$.
	Therefore, the global minimum of $(1+Div(p(z_{1})||p(z_{2})))^{-1} - \exp(-Div(p(z_{1})||p(z_{2})))$ is zero \\
	ii) $\lim_{\sigma_{2} \rightarrow \infty}(1+Div(p(z_{1})||p(z_{2}))) - \exp(Div(p(z_{1})||p(z_{2})))<0$. Therefore, $(1+Div(p(z_{1})||p(z_{2})))^{-1} - \exp(-Div(p(z_{1})||p(z_{2})))>0$ for sufficiently large $\sigma^{2}$.
	
\end{document}